\documentclass{article}

\PassOptionsToPackage{numbers}{natbib}
\usepackage[preprint]{neurips_2026}
\usepackage[T1]{fontenc}
\usepackage{hyperref}
\usepackage{url}
\usepackage{booktabs}
\usepackage{amsfonts}
\usepackage{amsmath}
\usepackage{nicefrac}
\usepackage{microtype}
\usepackage{float}
\usepackage[table]{xcolor}
\usepackage{graphicx}
\usepackage{adjustbox}
\usepackage{subcaption}
\usepackage[most]{tcolorbox}
\usepackage{xcolor}
\usepackage{listings}
\usepackage{caption}

\definecolor{promptblack}{RGB}{0,0,0}
\definecolor{promptgray}{RGB}{247,247,247}
\definecolor{tagblue}{RGB}{20,90,170}
\definecolor{tagpurple}{RGB}{130,60,160}
\definecolor{taggreen}{RGB}{30,130,80}
\definecolor{tagred}{RGB}{180,50,50}
\definecolor{ruleorange}{RGB}{180,100,20}

\lstdefinestyle{promptstyle}{
    basicstyle=\ttfamily\footnotesize,
    breaklines=true,
    breakatwhitespace=false,
    columns=fullflexible,
    keepspaces=true,
    showstringspaces=false,
    frame=none,
    backgroundcolor=\color{promptgray},
    moredelim=**[s][\color{tagblue}]{<}{>},
    moredelim=**[s][\color{tagpurple}]{[}{]},
    keywordstyle=\color{taggreen}\bfseries,
    keywords={Rules,Use,Follow,Please,Output,Then,Finally,If},
    alsoletter={_},
}

\newtcolorbox{promptbox}[1]{
    enhanced,
    colback=promptgray,
    colframe=black,
    boxrule=0.8pt,
    arc=1.5pt,
    left=6pt,
    right=6pt,
    top=6pt,
    bottom=6pt,
    title={#1},
    coltitle=white,
    fonttitle=\bfseries,
    colbacktitle=promptblack,
    attach boxed title to top left={
        xshift=0pt,
        yshift=-0.5pt
    },
    boxed title style={
        colback=promptblack,
        colframe=promptblack,
        boxrule=0pt,
        arc=1pt,
        left=6pt,
        right=6pt,
        top=3pt,
        bottom=3pt
    }
}

\newcommand{\ours}{\textsc{MinCU}}

\title{MinCU: A Fine-Grained Benchmark for Grounded Minimal-Change Understanding in Image Pairs}

\author{%
  \textbf{Chaoqian Mu}$^{1,*}$ \quad
  \textbf{Wenhao Wu}$^{1,*}$ \quad
  \textbf{Zichen Liang}$^{2}$ \quad
  \textbf{Jiaxu Li}$^{1}$ \\
  \textbf{Lijun Wang}$^{1}$ \quad
  \textbf{Yifan Wang}$^{1}$ \quad
  \textbf{Huchuan Lu}$^{1}$ \\[0.5em]
  {\normalfont $^{1}$Dalian University of Technology \quad $^{2}$Independent Researcher} \\[0.3em]
  {\normalfont\small\texttt{muchaoqian@mail.dlut.edu.cn} \quad \texttt{2021wwh@mail.dlut.edu.cn}} \\
  {\normalfont\small\texttt{liangzichen1009@163.com} \quad \texttt{liuhetongchun@mail.dlut.edu.cn}} \\
  {\normalfont\small\texttt{ljwang@dlut.edu.cn} \quad \texttt{wyfan@dlut.edu.cn} \quad \texttt{lhchuan@dlut.edu.cn}} \\[0.3em]
  {\normalfont\small $^{*}$Equal contribution.}
}

\hypersetup{
  pdftitle={MinCU: A Fine-Grained Benchmark for Grounded Minimal-Change Understanding in Image Pairs},
  pdfauthor={Chaoqian Mu, Wenhao Wu, Zichen Liang, Jiaxu Li, Lijun Wang, Yifan Wang, Huchuan Lu}
}

\begin{document}

\maketitle

\begin{abstract}
Localizing and describing fine-grained differences between near-identical images is a critical yet underexplored capability for multimodal large language models (MLLMs). Existing benchmarks largely assess semantic comparison or single-image grounding in isolation, without jointly requiring faithful description and physical localization. To bridge this gap, we introduce \ours, a benchmark for grounded minimal-change understanding, where each sample consists of an image pair differing by a single atomic variation in object category, attribute, count, or spatial position, and models are evaluated on their ability to describe the change, localize the changed regions, and identify the changed entity. We further propose Semantic-Guided Implicit Spatial Anchors (SG-ISA), a structured autoregressive method that decomposes prediction into a Think-Locate-Describe sequence. SG-ISA first predicts a semantic cue for the changed concept, then uses discrete spatial anchors as an implicit localization scaffold, and finally generates the change description together with the grounding box. Experiments reveal that even the strongest closed-source MLLMs and recent R1-style reasoning models struggle on \ours, with most failing to jointly produce accurate descriptions and grounding boxes. Compared to the previous chain-of-thought method, fine-tuning with SG-ISA yields substantial joint improvements in grounding accuracy and description quality while reducing reasoning-token overhead by approximately 26\%. These results suggest that an implicit intermediate spatial interface can be more effective than relying solely on model scale for grounded dual-image understanding.
\end{abstract}

\begin{figure}[htbp]
  \centering
  \includegraphics[width=\linewidth]{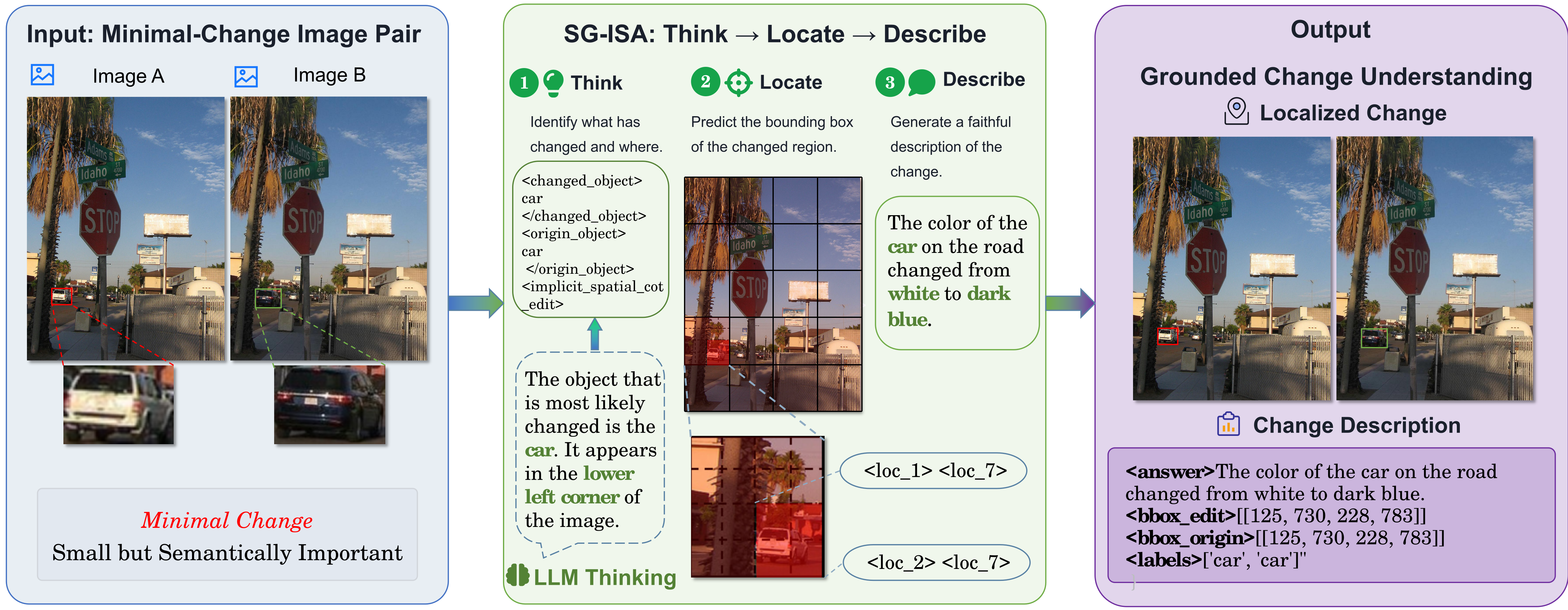}
  \caption{Overview of our \ours{} benchmark and the SG-ISA method. Given Image A and Image B differing by one localized atomic change (e.g., object, attribute, counting, or spatial relocation), a model must jointly describe the semantic difference, predict bounding boxes that localize the corresponding regions in both Image A and Image B, and identify the target entity.}
  \label{fig:task_overview}
  \vspace{-1.5em}
\end{figure}

\section{Introduction}

In recent years, fine-grained visual comparison has become an increasingly important capability for Multimodal Large Language Models (MLLMs), as many visual reasoning scenarios require identifying subtle local differences between near-identical images~\citep{radke2005image,bergmann2019mvtec,im2025longitudinal}. However, recognizing the presence of a visual difference alone is insufficient: a model should further explain \emph{what} has changed and ground the corresponding regions in both images. To study this capability, we formalize the task as \textbf{grounded minimal-change understanding}. Each example consists of two near-identical images, denoted as Image A and Image B, where Image B is synthesized from Image A with one localized atomic change during dataset construction. Under this formulation, a model is required to describe the visual difference and localize the corresponding regions in both Image A and Image B using spatial coordinates. This targets a fundamental requirement of real-world visual comparison: small, localized changes can substantially affect downstream interpretation, and thus must be both semantically understood and spatially grounded. However, most existing MLLMs are natively designed for single-image scenarios. Consequently, they are not directly equipped to localize minimal changes across near-identical image pairs, which requires complex cross-image comparison and reasoning over relative spatial differences.

Existing image-pair benchmarks provide important foundations for studying visual change understanding, but they only partially address the requirements of grounded minimal-change reasoning. Image Difference Captioning datasets, such as Spot-the-Diff~\citep{jhamtani2018learning}, CLEVR-Change~\citep{park_2019_robust}, OmniDiff~\citep{liu2025omnidiff}, and Image-Editing-Request~\citep{tan2019expressing}, focus on generating natural language descriptions of differences, without requiring explicit localization of the corresponding regions. Minimal-change benchmarks such as VisMin~\citep{awal2024vismin} further evaluate fine-grained semantic sensitivity, but are primarily formulated as caption-image matching tasks. More recently, DiffGround~\citep{wang2025diffground} introduced bounding-box annotations for changed regions, taking an important step toward grounded change understanding. However, existing grounded datasets largely emphasize spatially simple edits, such as in-place object replacement, where the changed and original regions remain closely aligned in position and scale. As a result, they provide limited coverage of minimal changes involving nontrivial cross-image spatial transformations, including object displacement, scale variation, and asymmetric one-to-many or many-to-one correspondences.

A similar gap exists on the modeling side. Coordinate-based grounding models, including Pix2Seq~\citep{chen2021pix2seq}, Shikra~\citep{chen2023shikra}, Ferret~\citep{you2024ferret}, Qwen3-VL~\citep{bai2025qwen3vl}, and InternVL3.5~\citep{wang2025internvl35}, have substantially advanced open-vocabulary visual grounding, but are primarily designed for single-image localization rather than cross-image change grounding. Consequently, they often predict target boxes without explicitly modeling the correspondence between the original and changed regions. Reasoning-oriented models such as Migician~\citep{li2025migician} and VLM-R1~\citep{shen2025vlmr1} introduce textual chain-of-thought to improve visual reasoning, but textual reasoning remains an indirect interface for two-dimensional spatial alignment. These limitations suggest the need for benchmarks and methods that explicitly evaluate whether models can jointly describe subtle changes and ground their corresponding regions across near-identical image pairs.

To address these specific challenges, we introduce \ours~(Grounded Minimal-Change Understanding), a rigorous benchmark and specialized training set designed specifically for fine-grained minimal visual change understanding and grounding. Meticulously constructed using a carefully devised data synthesis pipeline, \ours~comprises a large-scale instruction-tuning set of 8,000 image pairs and an evaluation suite of 2,000 test image pairs. Crucially, we systematically isolate four atomic minimal-change categories: object, attribute, counting, and relation, ensuring that models must handle both static appearance shifts and challenging spatial dynamism. Building on this newly-built testbed, we explore whether a compressed, geometrically bounded state can serve as a better intermediate representation for spatial reasoning than unconstrained text. To this end, we propose \textbf{Semantic-Guided Implicit Spatial Anchors (SG-ISA)}, an autoregressive framework that enforces a ``Think-Locate-Describe'' paradigm. By quantizing the 2D coordinate space into a 1D discrete spatial bottleneck, the model first emits a concise semantic cue, robustly locates the target region using a minimal set of discrete anchor tokens, and finally generates a spatially constrained description and precise bounding boxes. Extensive experiments on \ours~demonstrate that our approach achieves superior Grounded-F$_1$ performance. Furthermore, it reduces the computational token overhead of conventional Chain-of-Thought reasoning by approximately 26\%, enabling a 2B-parameter model to substantially outperform much larger generic multi-modal systems.

In summary, our contributions are summarized as follows:
\begin{itemize}
\item \textbf{A Novel Paradigm and Benchmark for Fine-Grained Minimal Visual Change Understanding:} We formalize the task of grounded minimal-change understanding, requiring models to jointly describe, localize, and semantically align fine-grained differences. To this end, we construct \ours, a rigorously controlled dual-image benchmark featuring dense bounding-box annotations across atomic change categories to systematically evaluate cross-image spatial reasoning.

\item \textbf{Token-Efficient Implicit Spatial Reasoning:} We propose Semantic-Guided Implicit Spatial Anchors (SG-ISA), an framework that structures generation into a ``Think-Locate-Describe'' sequence. By utilizing a quantized 1D spatial bottleneck instead of verbose natural language, SG-ISA achieves precise alignment while fundamentally eliminating the severe token overhead and latency associated with explicit textual CoT.

\item \textbf{Extensive Empirical Validation:} Experiments on \ours~demonstrate that SG-ISA substantially outperforms existing reasoning paradigms, enabling a lightweight 2B-parameter model to surpass much larger systems. Supported by comprehensive ablations, we systematically validate the proposed autoregressive decoding sequence for cross-image alignment, expose the limitations of direct coordinate regression, and demonstrate the impact of discrete anchor granularity on grounding precision.
\end{itemize}

\section{Related Work}

\subsection{Visual Grounding in Multimodal Large Language Models}

A recent line of work integrates spatial grounding into the autoregressive decoding of multimodal large language models (MLLMs). Earlier grounding and referring-expression systems also established the importance of single-image language-to-region alignment~\citep{kazemzadeh2014referitgame,mao2016generation,yu2016modeling,kamath2021mdetr,liu2023grounding}. Kosmos-2 \citep{peng2023kosmos2} represents bounding boxes as discrete location tokens and trains the model to ground referring expressions within the LLM vocabulary. Shikra \citep{chen2023shikra} demonstrates that a plain MLLM can handle coordinate inputs and outputs in natural-language form without extra modules. Ferret \citep{you2024ferret} introduces a hybrid region representation combining discrete coordinates with continuous features, enabling referring and grounding at arbitrary granularity. Qwen-VL \citep{bai2023qwen} and its successors \citep{bai2025qwen3vl} unify image-level understanding with region-level grounding in a single architecture. Beyond end-to-end models, Set-of-Mark prompting \citep{yang2023som} overlays visual marks on images to elicit grounding from frozen proprietary MLLMs such as GPT-4V~\citep{openai2023gpt4}. These methods demonstrate that MLLMs can effectively ground language in single images. However, they are not designed for the cross-image difference grounding scenario studied in this work, where the target region must be inferred by comparing two near-identical inputs.

\subsection{Grounding Chain-of-Thought and Multi-Image Reasoning}

Chain-of-thought (CoT)~\citep{wei2022chain} has been extended to multimodal settings to improve spatial and compositional understanding. Compositional CoT prompting~\citep{mitra2024compositional} decomposes complex visual queries into sub-problems for stepwise reasoning with MLLMs. Argus ~\citep{man2025argus} employs object-centric grounding as visual chain-of-thought signals, conditioning attention on intermediate spatial anchors. Grounded CoT~\citep{wu2025grounded} explicitly interleaves textual reasoning steps with spatial coordinate references. Several recent works target multi-image reasoning with explicit CoT. GeM-VG~\citep{zheng2026gem} proposes a generalized multi-image visual grounding framework using hybrid reinforcement learning with CoT reasoning. Migician ~\citep{li2025migician} introduces free-form multi-image grounding through end-to-end training. On the reinforcement learning side, R1-style reasoning models such as VLM-R1~\citep{shen2025vlmr1} and UniVG-R1~\citep{bai2025univgr1} train MLLMs via group-relative policy optimization (GRPO)~\citep{shao2024deepseekmath} to improve step-by-step grounding deliberation. However, these explicit textual reasoning approaches incur token overhead, with one-dimensional text traces serving as an indirect intermediate representation for tasks requiring precise 2D geometric alignment. 

\subsection{Image Difference Captioning}

Image difference captioning (IDC), also termed change captioning, aims to generate a natural-language description of the semantic differences between a pair of similar images. This task directly connects to our benchmark setting: both require the model to compare two near-identical inputs and verbalize the change. Early work on Spot-the-Diff~\cite{jhamtani2018learning} established an encoder-decoder baseline on surface-level web images paired with crowdsourced difference descriptions. CLEVR-Change~\cite{park_2019_robust} later introduced a synthetic benchmark that isolates semantic factors such as color, shape, and spatial position, enabling controlled evaluation of the sentence-level dynamics of visual comparison. Building on these foundations, several lines of research have improved IDC performance through instance-level fine-grained feature representations~\cite{huang2021idc}, pre-training with contrastive learning~\cite{yao2022pcl}, and CLIP-based contrastive fine-tuning for difference-aware encoding~\cite{guo2022clip4idc}. More recently, VIXEN~\cite{black2024vixen} jointly reasons over image pairs and text through a visual text comparison network, and OneDiff~\cite{hu2024onediff} unifies difference captioning across diverse domains with a generalist MLLM-based model. On the benchmark side, OmniDiff~\cite{liu2025omnidiff} curates a fine-grained benchmark spanning multiple change types, while DiffGround~\cite{wang2025diffground} bridges the gap between text and spatial coordinates by providing bounding-box annotations for difference specifications. Our work extends this direction by not only requiring faithful change descriptions but also demanding precise dual-image bounding-box grounding across four atomic change categories under a strictly controlled minimal-change regime.

\section{Benchmark Creation}
\label{sec:dataset}

We devise a pipeline to construct a grounded minimal-change image-pair benchmark, where each example contains an original image and an edited image with one localized semantic change. Given the two images, a model is required to produce a brief difference description, localize the changed region in the edited image, localize the corresponding region in the original image when it exists, and output semantic labels for the changed entity. The pipeline consists of three stages: Source Image Selection, where an LLM scores LVIS~\citep{gupta2019lvis} images to retain high-quality, semantically clear candidates; Edit Instruction Generation and Image Editing, where an LLM generates category-specific edit prompts for each selected image and Nano Banana 2~\citep{google2026nanobanana2} applies them to produce four edited variants; and Automatic Scoring and Human Verification, where an LLM evaluates instruction adherence and edit quality, followed by manual confirmation to discard failures.
Only image pairs passing these checks are included in the final dataset.

\begin{figure}[htbp]
  \centering
  \includegraphics[trim=37.5bp 94.5bp 15bp 76.5bp,clip,width=\linewidth]{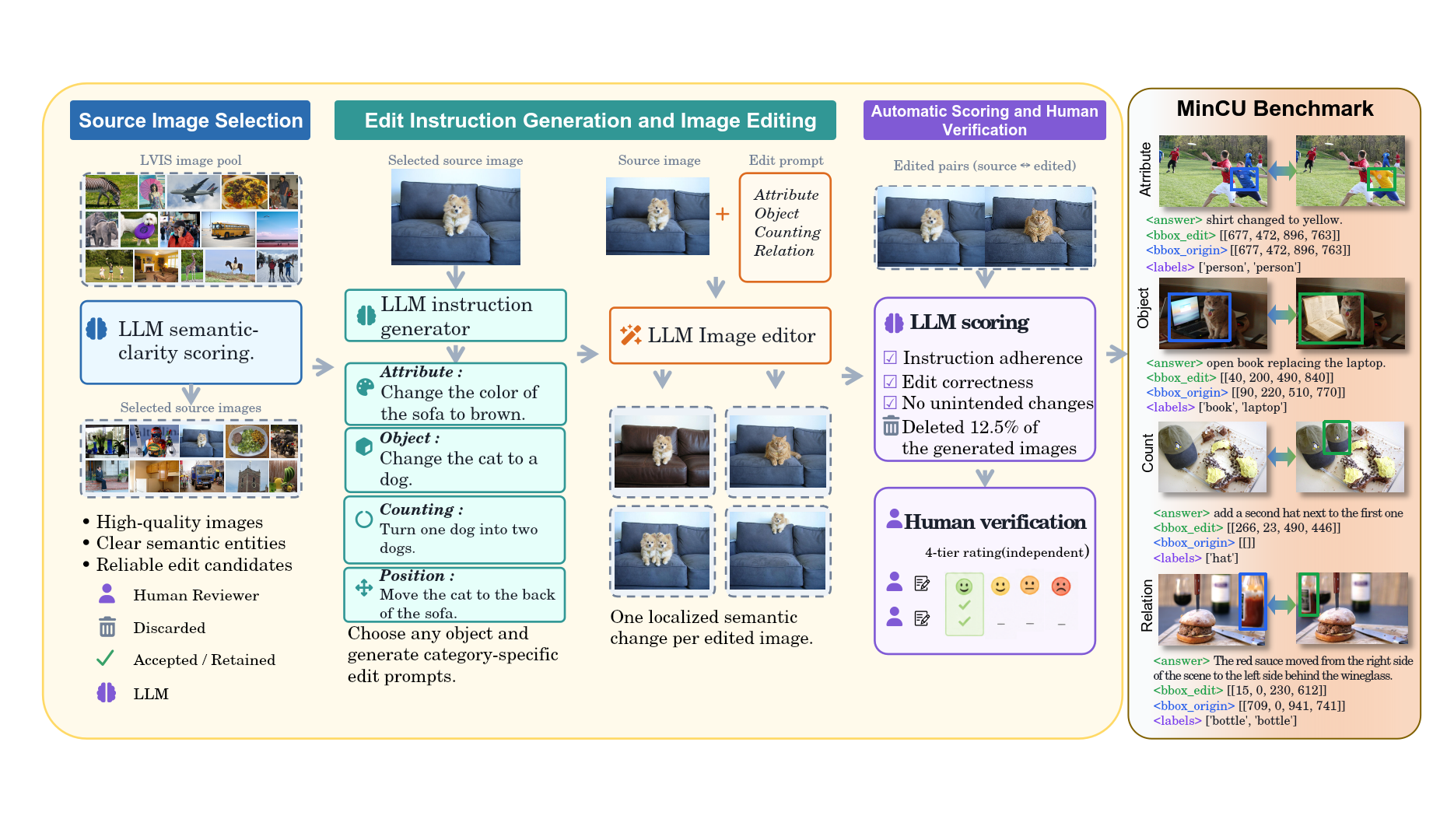}
  \caption{Three-stage \ours{} data pipeline: (1) Source image selection, where LVIS images are filtered for quality and semantic clarity; (2) Edit instruction generation, where category-specific prompts are created for each source image; Image editing, where Nano Banana~2 applies controlled local changes; and (3) Automatic scoring with human verification, where low-quality or ambiguous pairs are removed to produce the final benchmark.}
  \label{fig:pipeline}
\end{figure}

\noindent\textbf{Stage 1: Source image selection.}~
We start from images in the LVIS~\citep{gupta2019lvis} dataset. Since the benchmark targets fine-grained local changes, not all source images are equally suitable, they can make both editing and grounding unreliable. We therefore use a GPT-based scoring~\citep{openai2026gpt54} step to rank candidate images according to image quality and semantic clarity. Images with clearer object content and more reliable visual semantics are selected as source images for subsequent editing.

\noindent\textbf{Stage 2: Edit instruction generation and image editing.}~
For each selected source image, we generate category-specific edit instructions with Gemini~3.1~Pro Preview~\citep{google2026gemini31pro}, covering the four change types used in our benchmark: attribute, object, counting, and relation. These visually plausible and semantically unambiguous instructions are fed into Nano Banana~2~\citep{google2026nanobanana2} to produce four edited variants. The resulting ordered pairs are stored as $(I_A, I_B)$ and support precise difference description with explicit grounding annotation.

\noindent\textbf{Stage 3: Automatic scoring and human verification.}~
In total, we generate approximately 13,800 edited images. We first apply GPT-based scoring~\citep{openai2026gpt54} to filter out clear failures, roughly 12.5\% are discarded at this stage due to instruction non-adherence or unintended content changes. The remaining pairs then undergo dual-annotator manual verification, following the standard motivation of independent annotation for reliability control~\citep{artstein2008intercoder}. Two annotators independently rate each pair on a four-tier scale (excellent, good, fair, poor) and independently annotate grounding boxes. A pair is promoted to the benchmark only when both annotators assign it an excellent rating; this strict protocol removes a further $\sim$15\% of the surviving data. After verification, we select the 2,000 highest-quality pairs as the \ours{} benchmark and retain the remaining over 8,000 verified pairs as a training set.

\noindent\textbf{Benchmark composition.}~
The 2,000 curated pairs span four atomic change categories (500 each): \textbf{Attribute} (color, material, texture, shape, or state modifications), \textbf{Object} (replace, introduce, or remove an entity), \textbf{Counting} (alter the number of instances), and \textbf{Relation} (relocate an entity while preserving its identity). Images are tiered by the area ratio of the largest object: \textbf{Easy} ($>25\%$), \textbf{Medium} ($>10\%$ and $\le 25\%$), and \textbf{Hard} ($\le 10\%$), sampled at a $1:2:2$ ratio.

\section{Method}
\label{sec:method}

Existing multimodal large language models (MLLMs) typically approach dual-image difference localization through either direct regression in a continuous high-dimensional coordinate space or verbose explicit textual reasoning. To overcome the limitations of these unstructured or overly verbose paradigms, we propose Semantic-Guided Implicit Spatial Anchors (SG-ISA). SG-ISA is a structured autoregressive framework designed to inherently align semantic understanding with precise spatial reasoning through an implicit intermediate state.

\subsection{Task Formulation}

Given a pair of images A and B, denoted as $I_A$ and Image $I_B$ $\in \mathbb{R}^{H \times W \times 3}$, respectively, together with a language instruction $\mathcal{T}_{\mathrm{instruct}}$, our task is to produce structured outputs for difference identification, localization, and description. We decompose the target sequence into four ordered sub-sequences $\mathbf{Y} = [\mathbf{Y}_{\mathrm{obj}} \oplus \mathbf{Y}_{\mathrm{anchor}} \oplus \mathbf{Y}_{\mathrm{desc}} \oplus \mathbf{Y}_{\mathrm{box}}]$ to mimic human serialized cognition.

Unlike conventional end-to-end regression that directly fits the complex distribution $P(\mathbf{Y}_{\mathrm{box}} \mid \mathbf{X})$, SG-ISA decouples the black-box prediction into an implicit reasoning chain. Each generation step is constrained by prior results, where $\mathbf{Y}_{\mathrm{anchor}}$ acts as a pivotal latent variable to lower decoding entropy and compress the search space, echoing the information-bottleneck principle of retaining task-relevant compressed state. Benefiting from the cascaded generation paradigm, the entire SG-ISA framework supports standard end-to-end optimization via teacher forcing~\citep{williams1989learning,sutskever2014sequence}, with model parameters $\theta$ optimized by minimizing the negative log-likelihood of the sequential prediction:
\begin{equation}
\label{eq:loss}
\begin{aligned}
\mathcal{L}_{\mathrm{SFT}} 
&= -\mathbb{E}_{(\mathbf{X},\mathbf{Y})\sim\mathcal{D}} \Bigg[ \sum_{t=1}^{T} \log P_{\theta}\Big(y_t \,\big|\, y_{<t}, \mathbf{X}, 
\mathbf{Y}_{\mathrm{obj}} \oplus \mathbf{Y}_{\mathrm{anchor}} \oplus \mathbf{Y}_{\mathrm{desc}} \oplus \mathbf{Y}_{\mathrm{box}}\Big) \Bigg].
\end{aligned}
\end{equation}

\begin{figure}[htbp]
  \centering
  \includegraphics[width=0.96\linewidth]{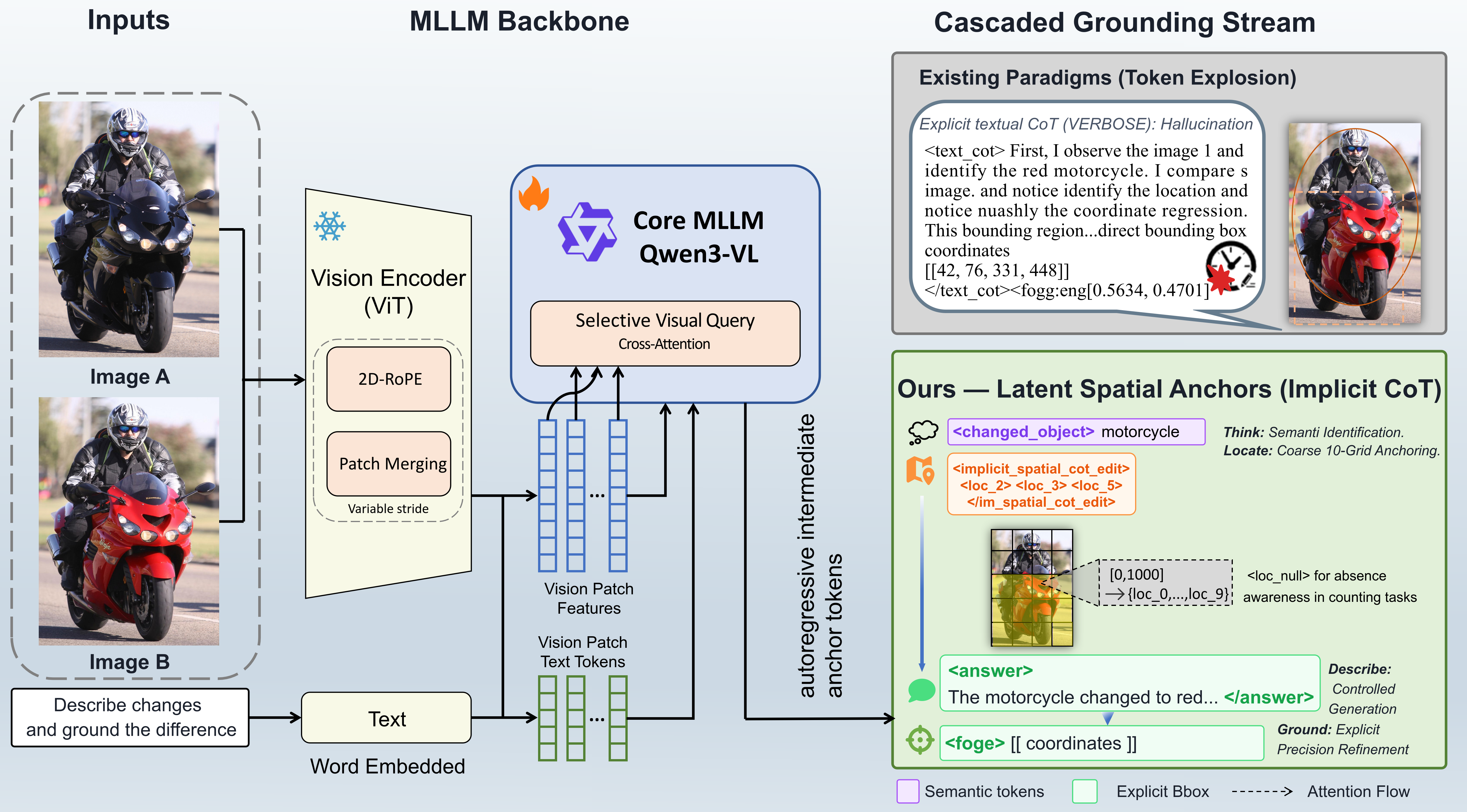}
  \caption{Overall methodology for SG-ISA. The joint probability distribution is factorized into Think-Locate-Describe stages, bridging the gap between semantic understanding and fine-grained localization via an implicit spatial bottleneck.}
  \label{fig:overall_methodology}
  \vspace{-1.5em}
\end{figure}

\subsection{The Think-Locate-Describe Paradigm}

Studies of human visual search suggest that attention and eye movements are guided by both visual and semantic information~\citep{wolfe1994guided,wu2014guidance}. To operationalize this ``scanning and gazing'' intuition, we structure the reasoning trajectory of SG-ISA into a cohesive ``Think-Locate-Describe'' pipeline. Rather than predicting free-form text and absolute coordinates in isolated or arbitrary orders, our framework intertwines them into a strictly ordered autoregressive stream. We detail each phase of this sequence below.

\textbf{Think: Semantic Hinting as Spatial Queries.} Human visual search is strongly shaped by semantic guidance in real-world scenes~\citep{wu2014guidance}. To emulate this cognitive prior, SG-ISA prompts the decoder to first generate a concise entity token sequence $\mathbf{Y}_{\mathrm{obj}}$. The contextualized hidden state of this semantic cue, denoted as $\mathbf{h}_{\mathrm{obj}} \in \mathbb{R}^{d}$, acts as a query vector. It interacts with the global visual representation $\mathbf{V}$ via cross-attention, defined as $\mathbf{A}_{\mathrm{vis}} = \mathrm{Softmax}((\mathbf{h}_{\mathrm{obj}} \mathbf{W}_Q)(\mathbf{V} \mathbf{W}_K)^{\top} / \sqrt{d_k})$. By emitting $\mathbf{Y}_{\mathrm{obj}}$ prior to any spatial prediction, the model can selectively activate the relevant visual feature patches in the latent space. This mechanism translates textual prompts into directed visual attention, establishing a semantic prior before coordinate prediction.

\textbf{Locate: Discrete Implicit Spatial Bottleneck.} Once triggered by a semantic cue, the human gaze performs a rapid saccade toward a coarse 2D neighborhood rather than calculating an absolute continuous boundary. Inspired by this fixation behavior, we substitute the traditional continuous coordinate search space with an implicit spatial sequence $\mathbf{Y}_{\mathrm{anchor}}$. Specifically, for a continuous bounding box $B = [x_1, y_1, x_2, y_2] \in [0, 1000]^4$, we define an element-wise quantization mapping $\mathcal{Q}(c) = \texttt{<loc\_i>}$, where $i = \min(\lfloor c/100 \rfloor, 9)$. This establishes a compact spatial vocabulary $\mathcal{V}_{\mathrm{loc}}$ of ten discrete tokens. Consequently, the model represents any altered region via a coarse 4-token sequence $\mathcal{Q}(B) = [\mathcal{Q}(x_1), \mathcal{Q}(y_1), \mathcal{Q}(x_2), \mathcal{Q}(y_2)]$, effectively mitigating spatial sparsity. These discrete grid tokens act as a compact \textit{information bottleneck}, filtering out complex background perturbations and geometrically constraining the multi-image alignment. 

\textbf{Describe: Spatially Constrained Generation.} Following semantic identification ($\mathbf{Y}_{\mathrm{obj}}$) and coarse spatial anchoring ($\mathbf{Y}_{\mathrm{anchor}}$), the decoder proceeds to generate the comprehensive natural language description $\mathbf{Y}_{\mathrm{desc}}$. At this stage, the autoregressive conditional probability distribution is mathematically updated to $P(\mathbf{Y}_{\mathrm{desc}} \mid \mathbf{Y}_{\mathrm{anchor}}, \mathbf{Y}_{\mathrm{obj}}, \mathbf{X})$. This structural dependence fundamentally mitigates a notorious flaw in MLLMs: the propensity of free-form text generation to ``wander off'' from the localized visual evidence. By conditioning the language model on the previously emitted spatial anchors, SG-ISA sharply reduces spatial-textual hallucinations. In the final decoding stage, the model directly outputs the precise bounding box coordinates $\mathbf{Y}_{\mathrm{box}} = [x_1, y_1, x_2, y_2]$ at a fine-grained 1000-bin resolution. The previously decoded coarse anchor tokens \texttt{<loc\_i>} naturally serve as a strong spatial prior within the context window, seamlessly bridging region-level approximation with exact coordinate regression without requiring complex post-processing.
\section{Experiments}

\subsection{Experimental Setup}

\textbf{Implementation Details \& Metrics.}
We fine-tune the Qwen3-VL-2B-Instruct backbone~\citep{bai2025qwen3vl} for 5 epochs via AdamW~\citep{loshchilov2017adamw} (LR=$10^{-5}$), updating the language model, connector, and 16 newly added implicit-anchor tokens while freezing the vision encoder (see Appendix~\ref{sec:hyperparameters} for full settings). The entire training process takes approximately two hours on 8 NVIDIA A800 (40GB) GPUs. 
We report mIoU, COCO-style mAP~\citep{lin2014coco}, and text quality using BLEU, ROUGE-L, METEOR, CIDEr, SPICE, BERTScore and SBERT similarity~\citep{papineni2002bleu,lin2004rouge,banerjee2005meteor,vedantam2015cider,anderson2016spice,zhang2019bertscore,reimers2019sbert}. Crucially, we introduce \textbf{Grounded F$_1$ (Gr-F$_1$)}, which registers a true positive only if spatial ($\text{IoU} \ge 0.5$), semantic ($\text{BERTScore} \ge 0.85$), and entity label constraints are simultaneously met.

\subsection{Main Results}

\textbf{Overall Performance.} We evaluate a broad spectrum of baselines: proprietary systems, open-source backbones, and specialized visual grounding models. Proprietary systems tested include GPT (GPT~5.4)~\citep{openai2026gpt54}, Gemini (Gemini~2.5~Pro, Gemini~3.1~Pro)~\citep{google2025gemini25pro,google2026gemini31pro}, Qwen3-VL-Plus~\citep{bai2025qwen3vl} and Doubao-seed-1.8~\citep{bytedanceseed2026seed18}. Open-source backbones include the Llama~3 family (Llama-3.2-11B)~\citep{grattafiori2024llama3}, InternVL~3.5 (8B/14B)~\citep{wang2025internvl35}, and the Qwen3-VL series (2B/4B/8B/30B-A3B). Specialized grounding models include VLM-R1~\citep{shen2025vlmr1}, UniVG-R1~\citep{bai2025univgr1}, and Migician~\citep{li2025migician}. Table~\ref{tab:main_results} suggests that strong single-image MLLM performance does not directly transfer to dual-image alignment. All proprietary and open-source baselines struggle with zero-shot grounding on our benchmark. Specialized frameworks also falter: Migician achieves only 47.76 Gr-F$_1$, while R1-style models (VLM-R1, UniVG-R1) degrade possibly because their training objectives are not tailored to cross-image change localization. In contrast, SG-ISA-enhanced Qwen3-VL-2B achieves 82.88 Gr-F$_1$ and 48.77 mAP, outperforming all baselines by a large margin across both localization and semantic-generation metrics.

\textbf{Fine-Grained Analysis.} Table~\ref{tab:change_type_grounding} and \ref{tab:change_type_map} report category-wise results (full breakdowns in Appendix Tables~\ref{tab:app_change_type_grounding} and~\ref{tab:app_change_type_semantic}). While object and attribute changes are relatively accessible, asymmetric tasks (Counting) and geometric shifts (Relation) remain formidable bottlenecks. The vanilla Qwen3-VL-2B SFT baseline suffers severe spatial confusion on these structural tasks (e.g., 46.86 Gr-F$_1$ on Counting). SG-ISA substantially resolves this, boosting Counting Gr-F$_1$ significantly. Notably, SG-ISA (2B) surpasses the much larger vanilla Qwen3-VL-Plus, suggesting that a principled spatial reasoning interface can be more important than sheer parametric scale for cross-image alignment in this setting.

\begin{table*}[t]
  \caption{Overall comparison on primary grounding and semantic metrics. Gr-F$_1$: Grounded F$_1$; HR: hallucination rate ($\downarrow$). Detailed overall results are deferred to Appendix Tables~\ref{tab:app_main_results_loc} and~\ref{tab:app_main_results_sem}.}
  \label{tab:main_results}
  \centering
  \footnotesize
  \setlength{\tabcolsep}{3pt}
  \begin{tabular}{lccccccc}
    \toprule
    Model & Gr-F$_1$ ($\uparrow$) & mIoU ($\uparrow$) & mAP ($\uparrow$) & HR ($\downarrow$) & CIDEr ($\uparrow$) & METEOR ($\uparrow$) & BERTScore ($\uparrow$) \\
    \midrule
    \multicolumn{8}{l}{\textit{Proprietary general-purpose MLLMs}} \\
    GPT~5.4~\citep{openai2026gpt54} & 19.58 & 24.35 & ~~1.96 & 90.00 & ~~3.93 & 18.87 & 89.12 \\
    Gemini~2.5~Pro~\citep{google2025gemini25pro} & 31.92 & 34.79 & ~~5.67 & 73.75 & ~~3.67 & 18.87 & 88.96 \\
    Gemini~3.1~Pro~\citep{google2026gemini31pro} & 42.89 & 44.97 & 27.98 & 71.55 & 10.07 & 23.12 & 90.69 \\
    Doubao-seed-1.8~\citep{bytedanceseed2026seed18} & 42.43 & 45.78 & 27.18 & 94.25 & ~~3.29 & 21.73 & 89.30 \\    
    Qwen3-VL-Plus~\citep{bai2025qwen3vl} & 48.43 & 56.01 & 35.69 & 79.75 & ~~0.67 & 26.99 & 87.01 \\
    \midrule
    \multicolumn{8}{l}{\textit{Open-source general-purpose MLLMs}} \\
    Llama-3.2-11B~\citep{grattafiori2024llama3} & ~~2.49 & ~~3.81 & ~~0.00 & 24.05 & ~~0.43 & ~~9.69 & 85.42 \\
    InternVL3.5-8B~\citep{wang2025internvl35} & 18.52 & 21.88 & ~~1.44 & 18.20 & ~~2.25 & 20.14 & 87.50 \\
    InternVL3.5-14B~\citep{wang2025internvl35} & 21.23 & 23.93 & ~~1.65 & 19.75 & ~~3.54 & 23.34 & 88.71 \\
    Qwen3-VL-2B~\citep{bai2025qwen3vl} & 13.26 & 20.30 & ~~3.04 & ~~7.95 & ~~0.00 & 18.75 & 85.17 \\
    Qwen3-VL-4B~\citep{bai2025qwen3vl} & 50.72 & 48.51 & 19.05 & 26.55 & ~~3.31 & 22.41 & 87.98 \\
    Qwen3-VL-8B~\citep{bai2025qwen3vl} & 48.84 & 47.27 & 16.09 & 23.60 & ~~6.05 & 25.70 & 88.17 \\
    Qwen3-VL-30B-A3B~\citep{bai2025qwen3vl} & 51.76 & 48.87 & 27.54 & 26.05 & ~~5.10 & 22.23 & 87.37 \\
    \midrule
    \multicolumn{8}{l}{\textit{Specialized visual-grounding models}} \\
    VLM-R1~\citep{shen2025vlmr1}(Zero-shot) & ~~0.29 & ~~0.76 & ~~0.12 & ~~\textbf{0.90} & ~~1.09 & ~~8.58 & 84.24 \\
    UniVG-R1~\citep{bai2025univgr1}(Zero-shot) & 14.11 & 13.84 & ~~3.53 & ~~2.90 & ~~0.79 & 12.93 & 84.13 \\
    Migician~\citep{li2025migician}(Zero-shot) & 47.76 & 46.16 & 20.14 & ~~9.45 & ~~4.30 & 22.35 & 85.67 \\
    \midrule
    \multicolumn{8}{l}{\textit{SG-ISA variants}} \\
    SFT Baseline (2B) & 72.75 & 64.41 & 38.10 & ~~7.25 & 18.72 & 43.83 & 91.94 \\
    \rowcolor{gray!10}
    SG-ISA~(2B) & 82.88 & 72.51 & 48.77 & ~~3.10 & 23.16 & 48.91 & 92.77 \\
    \rowcolor{gray!10}
    SG-ISA~(4B) & \textbf{83.17} & \underline{72.91} & \underline{49.94} & ~~3.05 & \underline{23.94} & \underline{49.38} & \underline{92.78} \\
    \rowcolor{gray!10}
    SG-ISA~(8B) & \underline{83.07} & \textbf{73.03} & \textbf{50.04} & ~~\underline{2.60} & \textbf{24.45} & \textbf{49.65} & \textbf{92.84} \\
    \bottomrule
  \end{tabular}
  \vspace{-1.5em} 
\end{table*}

\begin{table}[t]
  \begin{minipage}[t]{0.48\linewidth}
    \caption{Category-wise Grounded F$_1$ ($\uparrow$) on representative models. Full per-category localization metrics are given in Appendix Table~\ref{tab:app_change_type_grounding}.}
    \label{tab:change_type_grounding}
    \centering
    \footnotesize
    \setlength{\tabcolsep}{3pt}
    \resizebox{\linewidth}{!}{%
    \begin{tabular}{lccccc}
      \toprule
      Model & Object & Attr. & Rel. & Count. & Avg. \\
      \midrule
      GPT~5.4~\citep{openai2026gpt54} & 19.67 & 18.13 & 18.36 & 22.17 & 19.58 \\
      Gemini~2.5~Pro~\citep{google2025gemini25pro} & 32.68 & 27.27 & 35.34 & 32.40 & 31.92 \\
      Gemini~3.1~Pro~\citep{google2026gemini31pro} & 59.25 & 46.48 & 42.31 & 26.28 & 43.58 \\
      Doubao-seed-1.8~\citep{bytedanceseed2026seed18} & 44.82 & 41.31 & 40.85 & 42.81 & 42.43 \\
      Qwen3-VL-Plus~\citep{bai2025qwen3vl} & 67.64 & 58.95 & 48.21 & 21.96 & 49.19 \\
      Migician~\citep{li2025migician}(Zero-shot) & 59.06 & 67.54 & 43.51 & 17.59 & 47.76 \\
      SFT Baseline (2B) & 87.73 & 88.31 & 68.71 & 46.86 & 72.75 \\
      SG-ISA~(2B) & \underline{92.30} & \underline{90.56} & \textbf{81.74} & \underline{66.87} & \underline{82.88} \\
      SG-ISA~(8B) & \textbf{92.88} & \textbf{91.89} & \underline{80.37} & \textbf{67.05} & \textbf{83.07} \\
      \bottomrule
    \end{tabular}}
  \end{minipage}\hfill
  \begin{minipage}[t]{0.48\linewidth}
    \caption{Category-wise mAP ($\uparrow$) on representative models. Full per-category localization metrics are given in Appendix Table~\ref{tab:app_change_type_grounding}.}
    \label{tab:change_type_map}
    \centering
    \footnotesize
    \setlength{\tabcolsep}{3pt}
    \resizebox{\linewidth}{!}{%
    \begin{tabular}{lccccc}
      \toprule
      Model & Object & Attr. & Rel. & Count. & Avg. \\
      \midrule
      GPT~5.4~\citep{openai2026gpt54} & ~~2.26 & ~~1.57 & ~~1.99 & ~~2.65 & ~~1.96 \\
      Gemini~2.5~Pro~\citep{google2025gemini25pro} & ~~4.88 & 5.14 & ~~7.52 & ~~5.77 & ~~5.83 \\
      Gemini~3.1~Pro~\citep{google2026gemini31pro} & 38.13 & 37.86 & 27.95 & 16.45 & 30.10 \\
      Doubao-seed-1.8~\citep{bytedanceseed2026seed18} & 31.31 & 27.37 & 26.30 & 25.74 & 27.18 \\
      Qwen3-VL-Plus~\citep{bai2025qwen3vl} & 54.78 & 57.41 & 32.43 & 13.57 & 39.55 \\
      Migician~\citep{li2025migician}(Zero-shot) & 30.60 & 44.23 & 10.85 & ~~6.19 & 20.14 \\
      SFT Baseline (2B) & 58.51 & 60.81 & 33.52 & 13.74 & 38.10 \\
      SG-ISA~(2B) & \underline{64.33} & \underline{65.34} & \underline{46.33} & \underline{25.88} & \underline{48.77} \\
      SG-ISA~(8B) & \textbf{65.59} & \textbf{65.99} & \textbf{46.73} & \textbf{28.29} & \textbf{50.04} \\
      \bottomrule
    \end{tabular}}
  \end{minipage}
\end{table}

\begin{figure}[htbp]
  \centering
  \includegraphics[width=\linewidth]{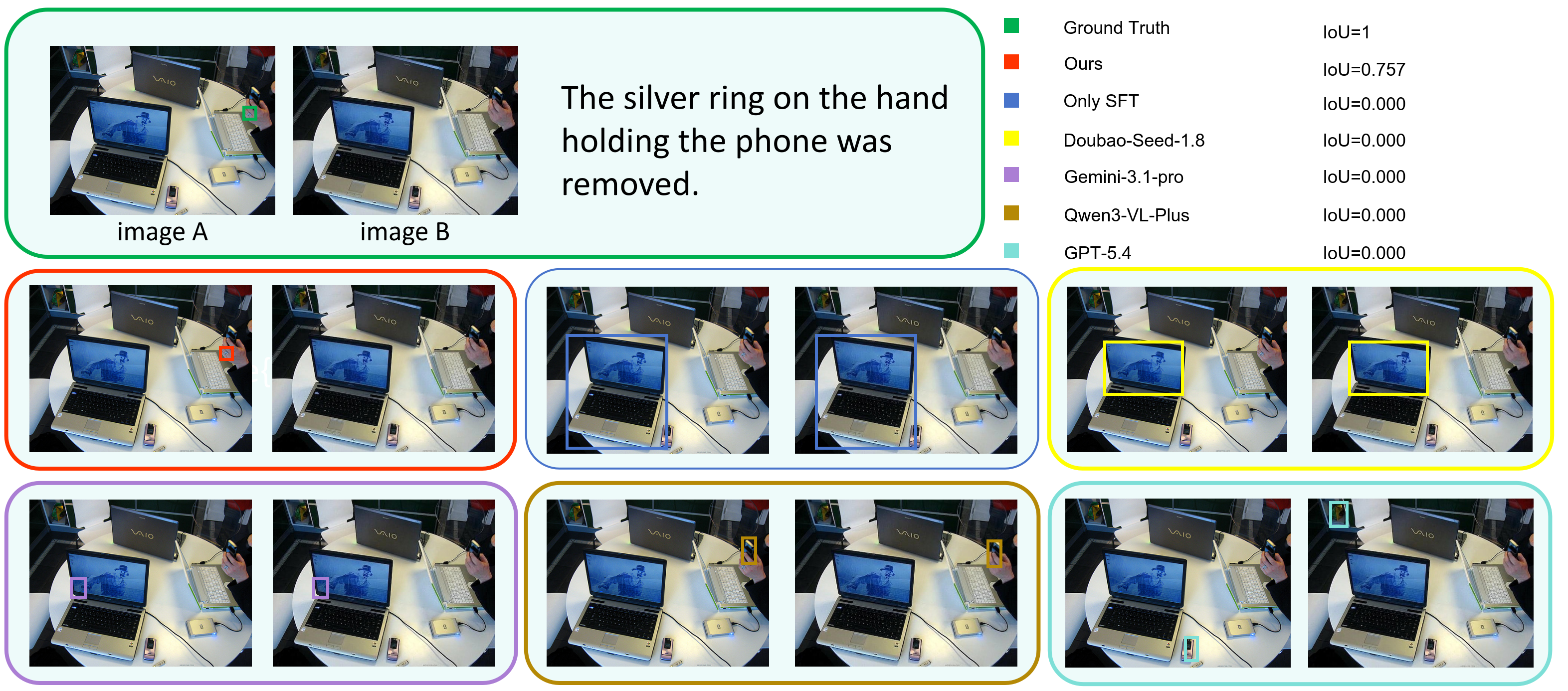}
  \caption{Qualitative comparison of SG-ISA against the SFT baseline and representative proprietary models. SG-ISA produces more precise bounding boxes and faithful descriptions.}
  \label{fig:qualitative_examples}
  \vspace{-1em}
\end{figure}

Figure~\ref{fig:qualitative_examples} presents representative qualitative comparisons between SG-ISA and strong baselines, including the SFT model and proprietary APIs. Relative to conventional decoding strategies, SG-ISA more reliably captures subtle local changes, handles asymmetric object appearance or disappearance, and avoids the localization and textual hallucinations that often arise when the model becomes spatially disoriented. Additional qualitative results are provided in Appendix Figure~\ref{fig:additional_vis_1} and~\ref{fig:additional_vis_2}. Furthermore, we demonstrate that SG-ISA's strong performance relies on genuine cross-image comparison rather than sequence memorization, maintaining high robustness under input sequence shuffling (see Appendix Table~\ref{tab:shuffle_results} and~\ref{tab:shuffle_category}). The model also explicitly exhibits robust zero-shot generalization out-of-the-box on in-the-wild real-world image pairs (see Appendix Table~\ref{tab:real_world_generalization}).

\subsection{Ablation Study}

\textbf{Decoding sequence exploration.}
We investigate the optimal arrangement of spatial anchors in the autoregressive generation pipeline. Table~\ref{tab:ablation_reason_stream} compares four variant structures. The autoregressive generation order has a substantial impact on the dual-image difference localization task. Forcing the model to output anchors prior to acquiring any semantic cues (Anchor-Before) appears less informed than first generating a semantic hint. Generating anchors after the complete natural language description (Anchor-After) still leaves room for error accumulation during long-sequence decoding. The proposed SG-ISA decoding sequence first activates the corresponding visual concepts via concise semantic hints, then constructs a spatial constraint framework with implicit anchors, so as to constrain the subsequent language generation and bounding box regression.

\textbf{Implicit vs.\ explicit CoT.}
As shown in Table~\ref{tab:ablation_reason_type}, we compare three reasoning strategies for multimodal tasks. The dominant explicit natural-language chain-of-thought (CoT) paradigm substantially outperforms the baseline, yet underperforms SG-ISA (Gr-F$_1$: 82.15 vs. 82.88) in tasks requiring strict 2D geometric alignment, demonstrating that 1D textual reasoning is suboptimal as an intermediate representation. In contrast, SG-ISA leverages a purely discrete implicit spatial CoT, which features a more compact and task-aligned representation: the average output token length is reduced from 135.77 to 100.58, and the maximum length drops from 512 to around 379.

\textbf{Granularity of implicit anchor.}
Table~\ref{tab:ablation_grid_granularity} studies the effect of coordinate discretization granularity. We compare the default 10-Grid design with a finer 100-Grid variant. The two settings are nearly tied in overall performance: 100-Grid obtains 83.04 Gr-F$_1$, while 10-Grid reaches 82.88. Although finer discretization offers higher nominal spatial resolution, it also enlarges the anchor vocabulary from 10 to 100 tokens and can introduce a more severe long-tail update problem during fine-tuning. For this reason, we adopt the 10-Grid configuration.

\begin{table}[!t]
\centering
\setlength{\tabcolsep}{3pt}
\small
\caption{\textbf{Ablation on decoding sequence.} We compare different autoregressive arrangements of the spatial bottleneck. Emitting the semantic cue (\texttt{[Obj]}) followed by spatial anchors (\texttt{[CoT]}) before generating the description yields the highest grounding accuracy and lower hallucination rate.}
\label{tab:ablation_reason_stream}
\begin{tabular}{lccccc}
\toprule
Variant & Generation stream & Gr-F$_1$ ($\uparrow$) & mIoU ($\uparrow$) & mAP ($\uparrow$) & HR ($\downarrow$) \\
\midrule
Baseline & [Ans]$\to$[Bbox] & 72.75 & 64.41 & 38.10 & 7.25 \\
Anchor-Before & [CoT]$\to$[Ans]$\to$[Bbox] & 79.47 & 70.16 & 45.36 & 6.60 \\
Anchor-After & [Ans]$\to$[CoT]$\to$[Bbox] & \underline{81.58} & \underline{71.53} & \underline{47.56} & \textbf{2.90} \\
\textbf{Ours} & \textbf{[Obj]$\to$[CoT]$\to$[Ans]$\to$[Bbox]} & \textbf{82.88} & \textbf{72.51} & \textbf{48.77} & \underline{3.10} \\
\bottomrule
\end{tabular}
\end{table}

\begin{table}[!t]
\centering
\setlength{\tabcolsep}{3pt}
\small
\caption{\textbf{Implicit vs.~explicit CoT.} Comparison of different reasoning modalities. Our discrete implicit spatial CoT achieves the highest localization precision (COCO-style mAP~\citep{lin2014coco}) while significantly reducing the decoding token overhead compared to explicit verbose reasoning.}
\label{tab:ablation_reason_type}
\begin{tabular}{lcccc}
\toprule
Reasoning type & Gr-F$_1$ ($\uparrow$) & mAP ($\uparrow$) & Avg. tokens & Max tokens \\
\midrule
None (Baseline) & 72.75 & 38.10 & \textbf{78.40} & \textbf{242} \\
Explicit Textual CoT & 82.15 & 48.21 & 135.77 & 512 \\
\textbf{Implicit Spatial CoT (Ours)} & \textbf{82.88} & \textbf{48.77} & 100.58 & 379 \\
\bottomrule
\end{tabular}
\end{table}

\begin{table}[!t]
\centering
\setlength{\tabcolsep}{3pt}
\small
\caption{\textbf{Impact of implicit spatial grid granularity.} We ablate the resolution of the intermediate coordinate vocabulary. The compressed 10-Grid configuration acts as an optimal information bottleneck, performing competitively with the 100-Grid while maintaining a more compact vocabulary.}
\label{tab:ablation_grid_granularity}
\begin{tabular}{lccccccccc}
\toprule
Grid size & Freq. & Gr-F$_1$ ($\uparrow$) & mIoU ($\uparrow$) & mAP ($\uparrow$) & CIDEr ($\uparrow$) & METEOR ($\uparrow$) \\
\midrule
Continuous (1000-bin) & Very sparse & 72.75 & 64.41 & 38.10 & 18.72 & 43.83  \\
Fine-grained (100-Grid) & Sparse & \textbf{83.04} & \textbf{72.59} & \underline{48.24} & \underline{21.16} & \underline{47.10} \\
Coarse-grained (10-Grid) & Dense & \underline{82.88} & \underline{72.51} & \textbf{48.77} & \textbf{23.16} & \textbf{48.91} \\
\bottomrule
\end{tabular}
\end{table}

\section{Conclusion}

In this work, we proposed \ours, a benchmark for grounded minimal-change understanding that jointly evaluates description fidelity and spatial localization across four atomic change categories. Extensive experiments reveal a substantial gap between existing MLLMs and the demands of fine-grained dual-image comparison: even the strongest proprietary and open-source models struggle to simultaneously describe and localize subtle changes. Based on this analysis, we introduced Semantic-Guided Implicit Spatial Anchors (SG-ISA), a structured autoregressive framework that enforces a Think-Locate-Describe decoding sequence. By decomposing the task into a semantic hint, a discrete implicit spatial bottleneck, and constrained description with coarse-to-fine bounding-box refinement, SG-ISA substantially improves both grounding accuracy and text quality while reducing reasoning-token overhead by approximately 26\%. Notably, a 2B-parameter model equipped with SG-ISA outperforms closed-source APIs by a wide margin, demonstrating that a principled spatial reasoning interface plays a more critical role than proprietary scale alone in our setting. Our benchmark and method establish a foundation for future work on cross-image spatial reasoning, with potential extensions to video change localization, medical longitudinal analysis, and fine-grained anomaly detection.

\bibliographystyle{plainnat}
\bibliography{references}

\appendix
\section*{Appendix}
\section{Broader Impacts}

Our work introduces \ours, a rigorous benchmark and the SG-ISA method for grounded minimal-change understanding in image pairs. By advancing the ability of Multimodal Large Language Models (MLLMs) to identify, describe, and accurately locate fine-grained visual differences, our research has substantial positive implications for domains requiring precision visual inspection, such as industrial defect detection and medical longitudinal imaging. Establishing precise spatial and semantic cross-image alignment can help mitigate visual hallucinations, fostering more reliable and accountable AI systems. Conversely, we acknowledge potential risks: systems capable of pinpointing subtle image shifts could theoretically be misused for localized surveillance or unintended adversarial tracking. We encourage the community to utilize our benchmark responsibly to develop transparent, robust, and aligned visual reasoning models.

\textbf{Ethical Considerations.} The \ours\ benchmark is constructed using publicly available images from the LVIS dataset, ensuring that no private or personally identifiable information (PII) is included. The paired images and instructional data are generated via accessible research models (Gemini 3.1 Pro and Nano Banana 2). The released data, model, and code will be provided under a research-only license that strictly prohibits malicious or unethical applications, such as unauthorized surveillance, disinformation, or privacy violations. We advocate for the responsible use of \ours\ to enhance safety and transparency in multimodal image comparison.

\section{Detailed Experiment Tables}

This section provides the full per-metric and per-category breakdowns of all evaluation results reported in the main paper. We evaluate localization quality using mIoU, COCO-style mAP~\citep{lin2014coco}, and our proposed Grounded F$_1$ (Gr-F$_1$; see Section~5.1 for definition). Text generation quality is measured by BLEU-4~\citep{papineni2002bleu}, ROUGE-L~\citep{lin2004rouge}, CIDEr~\citep{vedantam2015cider}, METEOR~\citep{banerjee2005meteor}, SPICE~\citep{anderson2016spice}, and BERTScore~\citep{zhang2019bertscore}.

\begin{table*}[!htbp]
  \caption{Detailed overall localization and joint metrics, including mean Intersection over Union (mIoU), COCO-style mean Average Precision (mAP)~\citep{lin2014coco} at multiple thresholds (AP@50, AP@75, AP@95), Grounded F$_1$ (Gr-F$_1$; see Section~5.1 for definition), and hallucination rate (HR; $\downarrow$).}
  \label{tab:app_main_results_loc}
  \centering
  \footnotesize
  \setlength{\tabcolsep}{3pt}
  \begin{tabular}{lccccccc}
    \toprule
    Model & Gr-F$_1$ ($\uparrow$) & mIoU ($\uparrow$) & mAP ($\uparrow$) & AP@50 ($\uparrow$) & AP@75 ($\uparrow$) & AP@95 ($\uparrow$) & HR ($\downarrow$) \\
    \midrule
    \multicolumn{8}{l}{\textit{Proprietary general-purpose MLLMs}} \\
    GPT~5.4~\citep{openai2026gpt54} & 19.58 & 24.35 & ~~1.96 & ~~7.65 & ~~1.22 & ~~0.02 & 90.00 \\
    Gemini~2.5~Pro~\citep{team2023gemini} & 31.92 & 34.79 & ~~5.67 & 14.81 & ~~4.94 & ~~0.13 & 73.75 \\
    Gemini~3.1~Pro~\citep{team2023gemini} & 42.89 & 44.97 & 27.98 & 40.07 & 30.13 & ~~6.52 & 71.55 \\
    Doubao-seed-1.8~\citep{bytedanceseed2026seed18} & 42.43 & 45.78 & 27.18 & 43.50 & 29.29 & ~~3.83 & 94.25 \\
    Qwen3-VL-Plus~\citep{bai2025qwen3vl} & 48.43 & 56.01 & 35.69 & 48.42 & 37.57 & ~~9.25 & 79.75 \\
    \midrule
    \multicolumn{8}{l}{\textit{Open-source general-purpose MLLMs}} \\
    Llama-3.2-11B~\citep{grattafiori2024llama3} & ~~2.49 & ~~3.81 & ~~0.00 & ~~0.00 & ~~0.00 & ~~0.00 & 24.05 \\
    InternVL3.5-8B~\citep{wang2025internvl35} & 18.52 & 21.88 & ~~1.44 & ~~4.20 & ~~1.01 & ~~0.03 & 18.20 \\
    InternVL3.5-14B~\citep{wang2025internvl35} & 21.23 & 23.93 & ~~1.65 & ~~5.10 & ~~1.25 & ~~0.00 & 19.75 \\
    Qwen3-VL-2B~\citep{bai2025qwen3vl} & 13.26 & 20.30 & ~~3.04 & ~~5.16 & ~~3.00 & ~~0.37 & ~~7.95\\
    Qwen3-VL-4B~\citep{bai2025qwen3vl} & 50.72 & 48.51 & 19.05 & 31.53 & 18.94 & ~~1.95 & 26.55 \\
    Qwen3-VL-8B~\citep{bai2025qwen3vl} & 48.84 & 47.27 & 16.09 & 30.00 & 15.78 & ~~0.94 & 23.60 \\
    Qwen3-VL-30B-A3B~\citep{bai2025qwen3vl} & 51.76 & 48.87 & 27.54 & 39.87 & 29.22 & ~~4.65 & 26.05 \\
    \midrule
    \multicolumn{8}{l}{\textit{Specialized visual-grounding models}} \\
    VLM-R1~\citep{shen2025vlmr1} & ~~0.29 & ~~0.76 & ~~0.12 & ~~0.99 & ~~0.03 & ~~0.00 & ~~0.90 \\
    UniVG-R1~\citep{bai2025univgr1} & 14.11 & 13.84 & ~~3.53 & ~~7.26 & ~~3.37 & ~~0.51 & ~~2.90 \\
    Migician~\citep{li2025migician} & 47.76 & 46.16 & 20.14 & 40.24 & 18.66 & ~~1.63 & ~~9.45 \\
    \midrule
    \multicolumn{8}{l}{\textit{SG-ISA variants}} \\
    SFT Baseline (2B) & 72.75 & 64.41 & 38.10 & 54.91 & 40.27 & ~~5.75 & ~~7.25 \\
    SG-ISA~(2B) & 82.88 & 72.51 & 48.77 & 69.62 & 51.13 & ~~8.99 & ~~3.10 \\
    SG-ISA~(4B) & 83.17 & 72.91 & 49.94 & 69.89 & 52.70 & ~~9.31 & ~~3.05 \\
    SG-ISA~(8B) & 83.07 & 73.03 & 50.04 & 69.92 & 52.83 & ~~9.50 & ~~2.60 \\
    \bottomrule
  \end{tabular}
\end{table*}

\begin{table*}[t]
  \caption{Detailed overall semantic-generation metrics ($\uparrow$), including BLEU-4~\citep{papineni2002bleu}, ROUGE-L~\citep{lin2004rouge}, CIDEr~\citep{vedantam2015cider}, METEOR~\citep{banerjee2005meteor}, SPICE~\citep{anderson2016spice}, and BERTScore~\citep{zhang2019bertscore}.}
  \label{tab:app_main_results_sem}
  \centering
  \footnotesize
  \setlength{\tabcolsep}{3pt}
  \begin{tabular}{lcccccc}
    \toprule
    Model & BLEU-4 & ROUGE-L & CIDEr & METEOR & SPICE & BERTScore \\
    \midrule
    \multicolumn{7}{l}{\textit{Proprietary general-purpose MLLMs}} \\
    GPT~5.4~\citep{openai2026gpt54} & ~~4.83 & 19.89 & ~~3.93 & 18.87 & 12.35 & 89.12 \\
    Gemini~2.5~Pro~\citep{team2023gemini} & ~~4.57 & 19.23 & ~~3.67 & 18.87 & 11.67 & 88.96 \\
    Gemini~3.1~Pro~\citep{team2023gemini} & ~~8.86 & 24.43 & 10.07 & 23.12 & 19.32 & 90.69 \\
    Doubao-seed-1.8~\citep{bytedanceseed2026seed18} & ~~5.73 & 22.01 & ~~3.29 & 21.73 & 13.32 & 89.30 \\
    Qwen3-VL-Plus~\citep{bai2025qwen3vl} & ~~3.08 & 19.96 & ~~0.67 & 26.99 & 14.67 & 87.01 \\
    \midrule
    \multicolumn{7}{l}{\textit{Open-source general-purpose MLLMs}} \\
    Llama-3.2-11B~\citep{grattafiori2024llama3} & ~~1.02 & ~~8.90 & ~~0.43 & ~~9.69 & ~~4.60 & 85.42 \\
    InternVL3.5-8B~\citep{wang2025internvl35} & ~~4.16 & 21.44 & ~~2.25 & 20.14 & 11.71 & 87.50 \\
    InternVL3.5-14B~\citep{wang2025internvl35} & ~~6.17 & 26.17 & ~~3.54 & 23.34 & 15.44 & 88.71 \\
    Qwen3-VL-2B~\citep{bai2025qwen3vl} & ~~1.84 & 13.29 & ~~0.00 & 18.75 & ~~0.00 & 85.17 \\
    Qwen3-VL-4B~\citep{bai2025qwen3vl} & ~~6.38 & 24.64 & ~~3.31 & 22.41 & 14.44 & 87.98 \\
    Qwen3-VL-8B~\citep{bai2025qwen3vl} & ~~9.34 & 28.78 & ~~6.05 & 25.70 & 19.32 & 88.17 \\
    Qwen3-VL-30B-A3B~\citep{bai2025qwen3vl} & ~~7.68 & 25.05 & ~~5.10 & 22.23 & 14.68 & 87.37 \\
    \midrule
    \multicolumn{7}{l}{\textit{Specialized visual-grounding models}} \\
    VLM-R1~\citep{shen2025vlmr1} & ~~0.53 & ~~9.35 & ~~1.09 & ~~8.58 & ~~3.84 & 84.24 \\
    UniVG-R1~\citep{bai2025univgr1} & ~~1.25 & 11.18 & ~~0.79 & 12.93 & ~~2.10 & 84.13 \\
    Migician~\citep{li2025migician} & ~~8.18 & 25.20 & ~~4.30 & 22.35 & 16.91 & 85.67 \\
    \midrule
    \multicolumn{7}{l}{\textit{SG-ISA variants}} \\
    SFT Baseline (2B) & 22.48 & 48.86 & 18.72 & 43.83 & 34.89 & 91.94 \\
    SG-ISA~(2B) & 27.98 & 53.12 & 23.16 & 48.91 & 39.87 & 92.77 \\
    SG-ISA~(4B) & 28.78 & 53.50 & 23.94 & 49.38 & 40.55 & 92.78 \\
    SG-ISA~(8B) & 28.82 & 53.76 & 24.45 & 49.65 & 40.50 & 92.84 \\
    \bottomrule
  \end{tabular}
\end{table*}

\begin{table*}[t]
  \caption{Detailed grounding metrics by change type ($\uparrow$), including Gr-F$_1$, mIoU, and COCO-style mAP~\citep{lin2014coco}.}
  \label{tab:app_change_type_grounding}
  \centering
  \scriptsize
  \setlength{\tabcolsep}{2pt}
  \begin{adjustbox}{width=\textwidth,center}
  \begin{tabular}{lcccccccccccc}
    \toprule
    \multicolumn{1}{c}{} & \multicolumn{3}{c}{Object} & \multicolumn{3}{c}{Attr.} & \multicolumn{3}{c}{Rel.} & \multicolumn{3}{c}{Count.} \\
    \cmidrule(lr){2-4} \cmidrule(lr){5-7} \cmidrule(lr){8-10} \cmidrule(lr){11-13}
    Model & Gr-F$_1$ & mIoU & mAP & Gr-F$_1$ & mIoU & mAP & Gr-F$_1$ & mIoU & mAP & Gr-F$_1$ & mIoU & mAP \\
    \midrule
    \multicolumn{13}{l}{\textit{Proprietary general-purpose MLLMs}} \\
    GPT~5.4~\citep{openai2026gpt54} & 19.67 & 22.58 & ~~2.26 & 18.13 & 21.36 & ~~1.57 & 18.36 & 21.41 & ~~1.99 & 22.17 & 23.43 & ~~2.65\\
    Gemini~2.5~Pro~\citep{google2025gemini25pro} & 21.64 & 26.29 & ~~2.98 & 18.98 & 26.49 & ~~3.12 & 26.04 & 29.00 & ~~5.02 & 22.24 & 25.99 & ~~3.47\\
    Gemini~3.1~Pro~\citep{google2026gemini31pro} & 59.25 & 65.48 & 38.13 & 46.48 & 68.43 & 37.86 & 42.31 & 55.21 & 27.95 & 26.28 & 29.68 & 16.45\\
    Doubao-seed-1.8~\citep{bytedanceseed2026seed18} & 44.82 & 52.01 & 31.31 & 41.31 & 47.67 & 27.37 & 40.85 & 46.74 & 26.30 & 42.81 & 50.48 & 25.74\\
    Qwen3-VL-Plus~\citep{bai2025qwen3vl} & 67.64 & 68.06 & 54.78 & 58.95 & 80.93 & 57.41 & 48.21 & 51.00 & 32.43 & 21.96 & 24.06 & 13.57\\
    \multicolumn{13}{l}{\textit{Open-source general-purpose MLLMs}} \\
    Llama-3.2-11B~\citep{grattafiori2024llama3} & ~~2.86 & ~~8.97 & ~~0.00 & ~~6.43 & 15.29 & ~~0.00 & ~~0.41 & ~~4.76 & ~~0.00 & ~~0.20 & ~~6.19 & ~~0.00\\
    Qwen3-VL-2B~\citep{bai2025qwen3vl} & 12.61 & 15.59 & ~~3.51 & 27.49 & 23.24 & 11.79 & ~~9.12 & ~~6.65 & ~~1.76 & ~~3.30 & 35.74 & ~~0.53\\
    InternVL3.5-8B~\citep{wang2025internvl35} & 20.79 & 23.81 & ~~2.43 & 27.60 & 30.27 & ~~2.91 & 14.56 & 12.02 & ~~0.76 & 10.83 & 23.54 & ~~0.81\\
    InternVL3.5-14B~\citep{wang2025internvl35} & 24.49 & 29.27 & ~~2.18 & 34.51 & 36.98 & ~~4.88 & 13.44 & 14.26 & ~~0.66 & 12.16 & 23.48 & ~~0.73\\
    Qwen3-VL-8B~\citep{bai2025qwen3vl} & 61.74 & 50.94 & 22.20 & 75.25 & 71.06 & 47.07 & 38.40 & 31.08 & ~~8.05 & 20.61 & 20.67 & ~~3.41\\
    Qwen3-VL-4B~\citep{bai2025qwen3vl} & 69.19 & 56.28 & 30.73 & 72.80 & 74.85 & 46.38 & 37.90 & 32.86 & 10.30 & 24.18 & 17.31 & ~~4.66\\
    Qwen3-VL-30B-A3B~\citep{bai2025qwen3vl} & 65.26 & 56.25 & 43.66 & 69.49 & 70.60 & 51.19 & 49.37 & 41.50 & 21.60 & 24.95 & 21.67 & ~~8.20\\
    \multicolumn{13}{l}{\textit{Specialized visual-grounding models}} \\
    VLM-R1~\citep{shen2025vlmr1} & ~~0.43 & ~~7.07 & ~~0.06 & ~~0.72 & ~~1.28 & ~~0.11 & ~~0.00 & ~~0.82 & ~~0.13 & ~~0.00 & 44.05 & ~~0.00\\
    UniVG-R1~\citep{bai2025univgr1} & 18.01 & 11.27 & ~~4.89 & 21.24 & ~~8.62 & ~~8.97 & 12.44 & ~~5.55 & ~~3.08 & ~~5.68 & 45.10 & ~~0.78\\
    Migician~\citep{li2025migician} & 59.06 & 60.51 & 30.60 & 67.54 & 73.95 & 44.23 & 43.51 & 42.81 & 10.85 & 17.59 & 28.64 & ~~6.19\\
    \multicolumn{13}{l}{\textit{SG-ISA variants}} \\
    SFT Baseline (2B) & 87.73 & 71.74 & 58.51 & 88.31 & 81.36 & 60.81 & 68.71 & 56.25 & 33.52 & 46.86 & 48.30 & 13.74\\
    SG-ISA~(2B) & 92.30 & 76.01 & 64.33 & 90.56 & 83.64 & 65.34 & 81.74 & 65.65 & 46.33 & 66.87 & 64.73 & 25.88\\
    SG-ISA~(4B) & 91.92 & 76.28 & 63.93 & 92.48 & 84.02 & 67.23 & 81.74 & 66.69 & 47.15 & 66.69 & 64.65 & 27.98\\
    SG-ISA~(8B) & 92.88 & 76.74 & 65.59 & 91.89 & 84.06 & 65.99 & 80.37 & 66.72 & 46.73 & 67.05 & 64.59 & 28.29\\
  \bottomrule
  \end{tabular}
  \end{adjustbox}
\end{table*}

\begin{table*}[htbp]
  \caption{Detailed semantic-generation metrics by change type ($\uparrow$), including CIDEr~\citep{vedantam2015cider}, METEOR~\citep{banerjee2005meteor}, and BERTScore~\citep{zhang2019bertscore}.}
  \label{tab:app_change_type_semantic}
  \centering
  \scriptsize
  \setlength{\tabcolsep}{2pt}
  \begin{adjustbox}{width=\textwidth,center}
  \begin{tabular}{lcccccccccccc}
    \toprule
    \multicolumn{1}{c}{} & \multicolumn{3}{c}{Object} & \multicolumn{3}{c}{Attr.} & \multicolumn{3}{c}{Rel.} & \multicolumn{3}{c}{Count.} \\
    \cmidrule(lr){2-4} \cmidrule(lr){5-7} \cmidrule(lr){8-10} \cmidrule(lr){11-13}
    Model & CIDEr & METEOR & BERT & CIDEr & METEOR & BERT & CIDEr & METEOR & BERT & CIDEr & METEOR & BERT \\
    \midrule
    \multicolumn{13}{l}{\textit{Proprietary general-purpose MLLMs}} \\
    GPT~5.4~\citep{openai2026gpt54} & ~~3.97 & 31.26 & 89.10 & ~~4.75 & 33.20 & 89.18 & ~~3.46 & 31.86 & 88.99 & ~~4.12 & 32.54 & 89.19\\
    Gemini~2.5~Pro~\citep{google2025gemini25pro} & ~~3.50 & 33.88 & 88.86 & ~~3.68 & 34.88 & 88.99 & ~~4.29 & 34.47 & 89.15 & ~~3.80 & 34.11 & 88.85\\
    Gemini~3.1~Pro~\citep{google2026gemini31pro} & ~~11.31 & 38.41 & 91.22 & 13.76 & 45.90 & 91.94 & ~~7.93 & 33.51 & 90.52 & ~~6.12 & 30.73 & 89.08\\
    Doubao-seed-1.8~\citep{bytedanceseed2026seed18} & ~~3.95 & 36.30 & 89.29 & ~~3.51 & 34.83 & 89.40 & ~~3.30 & 34.61 & 89.24 & ~~2.96 & 34.02 & 89.27\\
    Qwen3-VL-Plus~\citep{bai2025qwen3vl} & ~~0.81 & 31.51 & 87.60 & ~~0.12 & 28.90 & 87.50 & ~~1.27 & 25.56 & 87.12 & ~~0.56 & 21.97 & 85.84\\
    \multicolumn{13}{l}{\textit{Open-source general-purpose MLLMs}} \\
    Llama-3.2-11B~\citep{grattafiori2024llama3} & ~~0.39 & 14.50 & 85.46 & ~~0.44 & 15.25 & 85.47 & ~~0.67 & 17.58 & 86.10 & ~~0.21 & 13.48 & 84.64\\
    Qwen3-VL-2B~\citep{bai2025qwen3vl} & ~~0.00 & 19.95 & 85.24 & ~~0.00 & 20.96 & 85.71 & ~~0.00 & 18.89 & 85.48 & ~~0.00 & 15.21 & 84.22\\
    InternVL3.5-8B~\citep{wang2025internvl35} & ~~2.10 & 21.79 & 87.43 & ~~4.42 & 33.07 & 89.07 & ~~1.22 & 21.40 & 87.23 & ~~0.79 & 18.84 & 86.27\\
    InternVL3.5-14B~\citep{wang2025internvl35} & ~~3.32 & 26.74 & 88.69 & ~~6.59 & 36.52 & 90.30 & ~~2.03 & 25.74 & 88.56 & ~~1.71 & 23.09 & 87.30\\
    Qwen3-VL-8B~\citep{bai2025qwen3vl} & ~~7.81 & 33.33 & 88.68 & ~~8.55 & 37.98 & 89.28 & ~~4.60 & 26.07 & 88.09 & ~~2.81 & 24.82 & 86.62\\
    Qwen3-VL-4B~\citep{bai2025qwen3vl} & ~~3.88 & 31.39 & 88.65 & ~~6.57 & 34.85 & 89.25 & ~~1.62 & 25.03 & 87.67 & ~~0.89 & 22.52 & 86.34\\
    Qwen3-VL-30B-A3B~\citep{bai2025qwen3vl} & ~~6.42 & 30.54 & 87.68 & ~~5.66 & 34.42 & 87.97 & ~~3.63 & 26.00 & 87.35 & ~~3.98 & 25.50 & 86.45\\
    \multicolumn{13}{l}{\textit{Specialized visual-grounding models}} \\
    VLM-R1~\citep{shen2025vlmr1} & ~~1.40 & 13.08 & 84.50 & ~~1.34 & 13.04 & 84.48 & ~~1.07 & 12.31 & 84.29 & ~~0.83 & 10.56 & 83.67\\
    UniVG-R1~\citep{bai2025univgr1} & ~~0.70 & ~~7.32 & 83.85 & ~~1.23 & 10.81 & 84.11 & ~~0.69 & ~~9.04 & 84.33 & ~~0.44 & ~~7.39 & 84.25\\
    Migician~\citep{li2025migician} & ~~5.86 & 20.73 & 86.19 & ~~5.22 & 15.50 & 86.12 & ~~2.23 & 12.21 & 85.97 & ~~3.99 & 16.02 & 84.57\\
    \multicolumn{13}{l}{\textit{SG-ISA variants}} \\
    SFT Baseline (2B) & 21.08 & 47.32 & 92.37 & 21.84 & 49.96 & 93.10 & 17.46 & 44.92 & 92.07 & 13.15 & 33.13 & 90.23\\
    SG-ISA~(2B) & 24.94 & 52.84 & 92.90 & 24.59 & 52.35 & 93.42 & 22.74 & 49.65 & 93.04 & 19.37 & 40.76 & 91.68\\
    SG-ISA~(4B) & 25.32 & 53.14 & 93.00 & 26.20 & 53.21 & 93.67 & 23.79 & 50.20 & 92.99 & 19.44 & 40.93 & 91.45\\
    SG-ISA~(8B) & 26.63 & 53.85 & 93.03 & 26.71 & 53.23 & 93.51 & 24.04 & 50.37 & 93.00 & 19.68 & 41.11 & 91.83\\
  \bottomrule
  \end{tabular}
  \end{adjustbox}
\end{table*}

\section{Hyperparameters}
\label{sec:hyperparameters}

All training experiments are conducted on 8 NVIDIA A800 (40GB) GPUs. The complete fine-tuning process spans 5 epochs and takes approximately two hours. We employ the AdamW optimizer with a gradient accumulation step size of 4. The initial learning rate is set to $1\times 10^{-5}$, governed by a cosine annealing learning rate schedule. Additionally, the maximum sequence length is bounded at 4096, and we utilize DeepSpeed ZeRO Stage-2 (ZeRO-2) optimization to efficiently manage memory footprint during training.

\section{Ablation Study on \texttt{<loc\_null>} Token}
\label{sec:appendix_locnull_ablation}

We conduct an ablation study to evaluate the impact of the \texttt{<loc\_null>} token on the localization performance of our model, with all experiments conducted on the Counting split of the benchmark. The results are presented in Table~\ref{tab:ablation_locnull}.

\begin{table}[!htbp]
\centering
\setlength{\tabcolsep}{4pt} %
\small
\caption{Ablation study on the \texttt{<loc\_null>} token. All metrics are reported on the Counting split, with higher values indicating better performance for all localization metrics ($\uparrow$). mAP denotes COCO-style mean Average Precision~\citep{lin2014coco}.}
\label{tab:ablation_locnull}
\begin{tabular}{lccccccc}
\toprule
Variant & Gr-F$_1$ ($\uparrow$) & mIoU ($\uparrow$) & mAP ($\uparrow$) & AP@50 & AP@75 & AP@95 \\
\midrule
Baseline & 46.86 & 48.30 & 13.74 & 23.77 & 13.44 & 1.26 \\
w/o \texttt{<loc\_null>} & 65.83 & 63.84 & \textbf{26.45} & \textbf{45.22} & \textbf{26.19} & 2.55 \\
\textbf{w/ \texttt{<loc\_null>}} & \textbf{66.87} & \textbf{64.73} & 25.88 & 44.49 & 25.30 & \textbf{2.64} \\
\bottomrule
\end{tabular}
\end{table}

The inclusion of the \texttt{<loc\_null>} token yields consistent improvements in core localization metrics: Gr-F$_1$ increases by 1.04 points (from 65.83 to 66.87) and mIoU rises by 0.89 points (from 63.84 to 64.73). While mAP and AP@50 show minor decreases, AP@95 (a stricter metric for precise localization) improves from 2.55 to 2.64, demonstrating that \texttt{<loc\_null>} enhances the model's ability to produce spatially accurate predictions for counting-related tasks. This token effectively mitigates spatial hallucinations by explicitly modeling regions with no countable changes, leading to more robust localization performance overall.

\section{Structured vs.\ Free-Form Prediction Comparison}
\label{sec:appendix_structured_freeform}

We compare the performance of our structured SG-ISA (2B) model against the vanilla Qwen3-VL-2B-Instruct baseline using the same ground-truth annotations, evaluating both models' language outputs across standard captioning metrics including BLEU-4~\citep{papineni2002bleu}, ROUGE-L~\citep{lin2004rouge}, CIDEr~\citep{vedantam2015cider}, METEOR~\citep{banerjee2005meteor}, BERTScore~\citep{zhang2019bertscore}, and SBERT cosine similarity~\citep{reimers2019sbert}. The results are shown in Table~\ref{tab:structured_vs_freeform}.

\begin{table}[ht]
\centering
\caption{Results comparison between SG-ISA (structured) and Qwen3-VL-2B-Instruct (free-form) predictions under the same ground-truth annotations.}
\label{tab:structured_vs_freeform}
\begin{tabular}{lcc}
\toprule
Metric & SG-ISA (2B) & Qwen3-VL-2B-Instruct\\
\midrule
BLEU-4 & 1.15 & 0.37 \\
METEOR & 8.42 & 7.64 \\
ROUGE-L & 16.88 & 7.33 \\
CIDEr & 20.96 & 0.00 \\
BERTScore & 85.06 & 82.31 \\
SBERT cosine similarity & 32.31 & 31.75 \\
\bottomrule
\end{tabular}
\end{table}

The structured generation paradigm adopted by SG-ISA consistently outperforms the free-form baseline across all metrics. Notably, SG-ISA achieves substantial gains in sequence-level metrics such as BLEU-4 ($+0.78$), ROUGE-L ($+9.55$), and CIDEr ($+20.96$), highlighting its superior ability to align with the structured nature of the task description. Even on semantic similarity metrics that are less sensitive to surface form, SG-ISA maintains higher scores (BERTScore F1: $85.06$ vs.\ $82.31$; SBERT cosine: $32.31$ vs.\ $31.75$), indicating that structured generation not only improves format compliance but also preserves semantic fidelity better than unconstrained free-form outputs. The near-zero CIDEr score for the free-form baseline further suggests that open-ended generation struggles to produce descriptions that match the reference style and content distribution required by the task.

\section{Special Token Definitions}
\label{sec:app_token_details}

To support implicit spatial reasoning and coordinate quantization, we extend the tokenizer vocabulary with the following 16 special tokens:
\begin{itemize}
  \item \texttt{<loc\_0>} through \texttt{<loc\_9>} — 10 discrete coordinate bins for coarse spatial anchoring;
  \item \texttt{<loc\_null>} — empty-coordinate placeholder for asymmetric changes (object addition/removal);
  \item \texttt{<box\_sep>} — separator between individual bounding-box coordinates;
  \item \texttt{<implicit\_spatial\_cot\_origin>} and \texttt{</implicit\_spatial\_cot\_origin>} — markers enclosing the implicit spatial chain-of-thought for the original image;
  \item \texttt{<implicit\_spatial\_cot\_edit>} and \texttt{</implicit\_spatial\_cot\_edit>} — markers enclosing the implicit spatial chain-of-thought for the edited image.
\end{itemize}

\begin{figure}[!htbp]
\centering
\begin{promptbox}{Prompt Template for SG-ISA}
\begin{lstlisting}[style=promptstyle]
<image><image>
These two pictures are very similar. Please observe carefully: what changes have occurred in the second picture compared to the first one? Please provide a brief description of the changes.

Follow the human visual cognition process (Think-Locate-Describe-Ground): first identify what changed, then roughly locate both the new and original regions, then describe the change, and finally pinpoint the exact boundaries.

Use the following response format:
<changed_object> changed object name </changed_object>
<origin_object> original object name </origin_object>
<implicit_spatial_cot_edit> edit coarse anchors </implicit_spatial_cot_edit>
<implicit_spatial_cot_origin> origin coarse anchors </implicit_spatial_cot_origin>
<answer> brief description of changes
<bbox_edit>[[x1, y1, x2, y2]]
<bbox_origin>[[x1, y1, x2, y2]]
<labels>['label_name']

Rules:
1. Output <changed_object> first, i.e., the object in the edited image, then <origin_object>, i.e., the object in the original image.
2. Then output coarse spatial anchors for the changed region using <loc_*> tokens in x1 y1 x2 y2 order.
3. Then output coarse spatial anchors for the original region using <loc_*> tokens in x1 y1 x2 y2 order.
4. Then provide the full natural language description of the change.
5. Finally output the precise bounding boxes for both edit and origin regions.
6. If there are multiple boxes, separate them with <box_sep>.
7. If a side has no boxes, output <loc_null> for coarse anchors and [] for bbox.
\end{lstlisting}
\end{promptbox}
\caption{Prompt template used for structured autoregressive reasoning and grounding.}
\label{fig:prompt-template}
\end{figure}
\begin{figure}[!htbp]
\centering
\begin{promptbox}{Prompt Template for Edit Instruction Generation}
\begin{lstlisting}[style=promptstyle]
This is an image of a real-world scene. 
The following objects are officially annotated in this image: [{obj_list_str}]. 
It can undergo four types of changes: Objects, Attributes, Quantity, and Spatial relationships. 

CRITICAL REQUIREMENTS:
1. You MUST generate EXACTLY FOUR changes.
2. You MUST use EACH of the four change types exactly once (one Objects, one Attributes, one Quantity, one Spatial relationships). No duplicates!
3. Please select ONE OR MORE objects STRICTLY from the provided list ({obj_list_str}).
4. You MUST describe the change using the EXACT markdown format below:

1. Change Type: [Objects / Attributes / Quantity / Spatial relationships]
* Object before change: [Name of the original object from the list]
* Object after change: [Name of the modified object / New state]
* Content of Change: [A concise prompt describing the action]
* Resulting State: [Description of the final scene]

(Repeat this structure for items 2, 3, and 4)
\end{lstlisting}
\end{promptbox}
\caption{Prompt template used for edit instruction generation.}
\label{fig:prompt-template_2}
\end{figure}
\section{Failure Cases and Limitations}

\begin{figure}[htbp]
    \centering
    \begin{subfigure}[b]{0.48\linewidth}
        \centering
        \includegraphics[width=\linewidth]{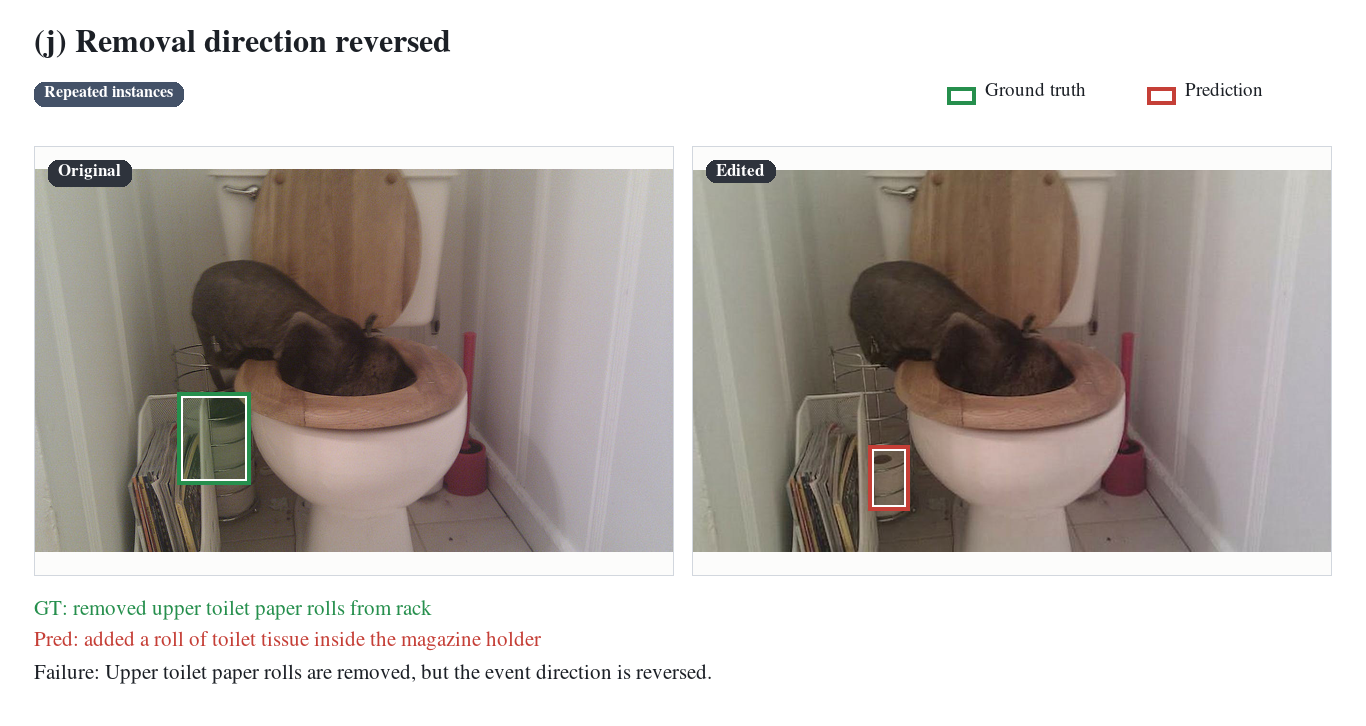}
    \end{subfigure}
    \hfill
    \begin{subfigure}[b]{0.48\linewidth}
        \centering
        \includegraphics[width=\linewidth]{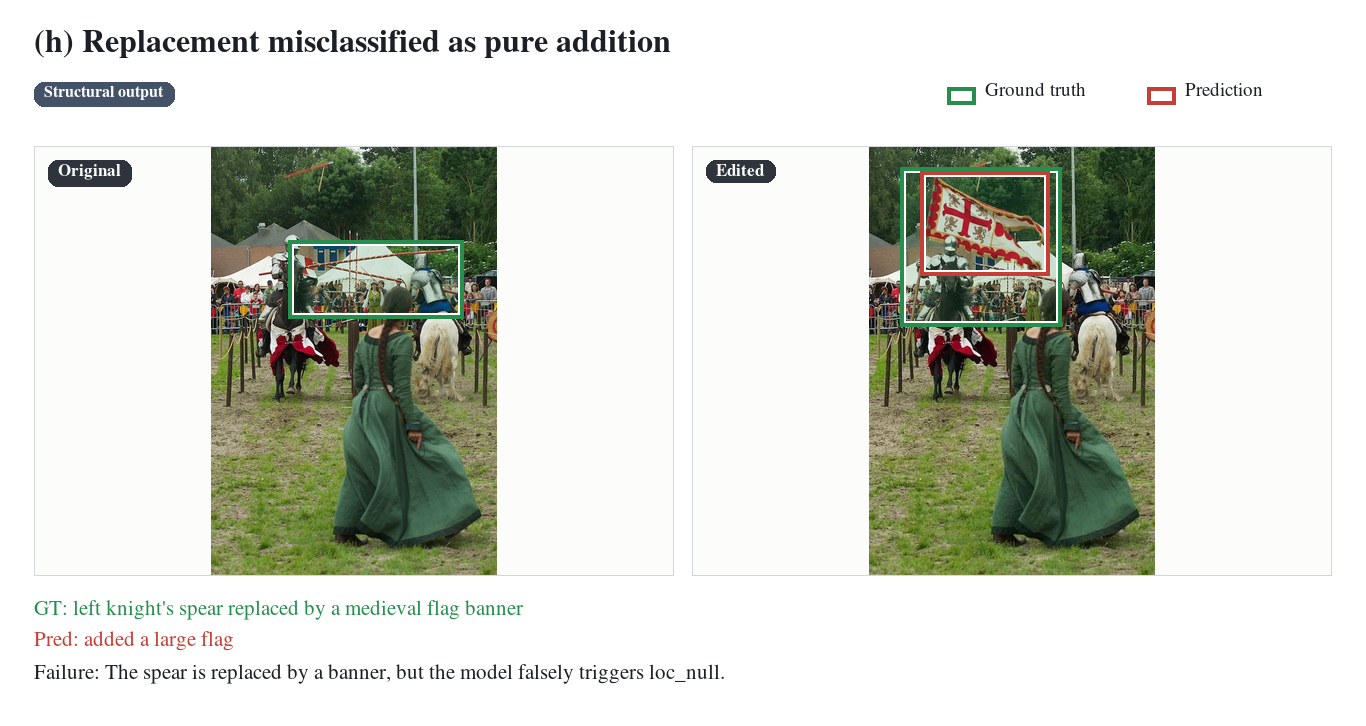}
    \end{subfigure}
    
    \vspace{0.3em}
    \begin{subfigure}[b]{0.48\linewidth}
        \centering
        \includegraphics[width=\linewidth]{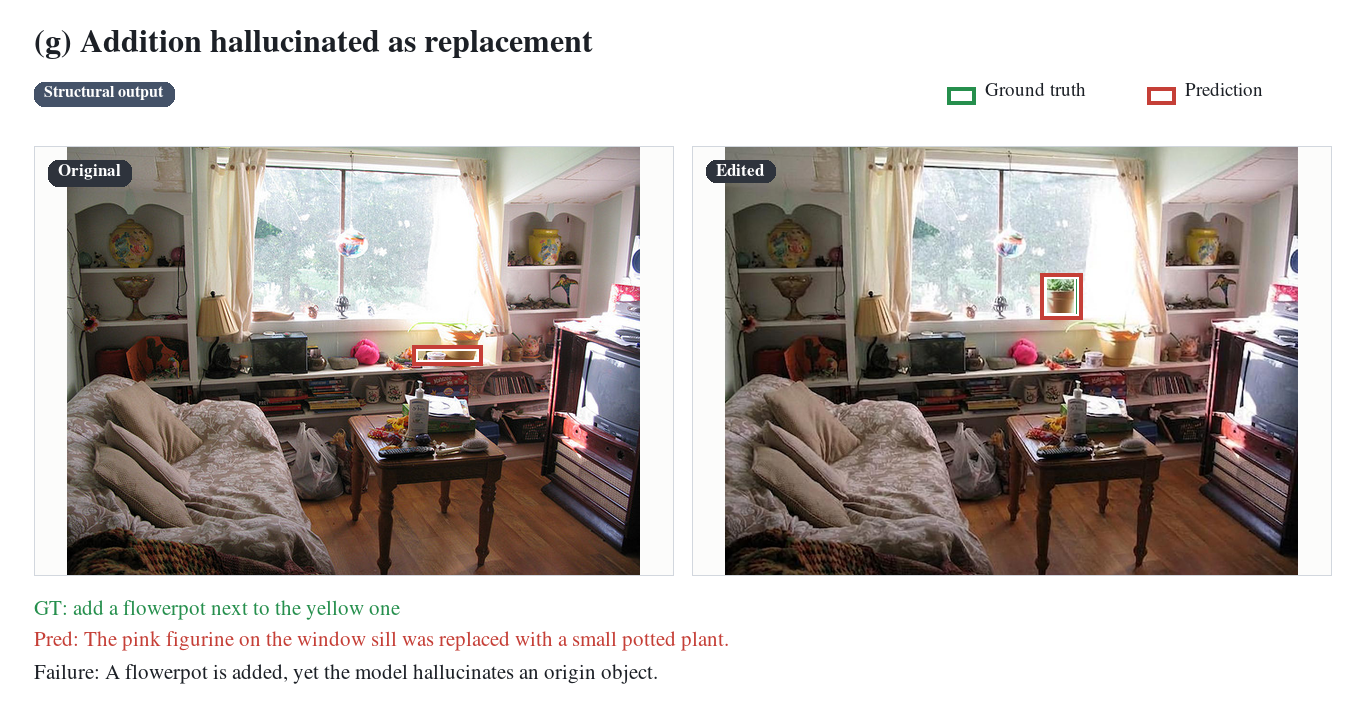}
    \end{subfigure}
    \hfill
    \begin{subfigure}[b]{0.48\linewidth}
        \centering
        \includegraphics[width=\linewidth]{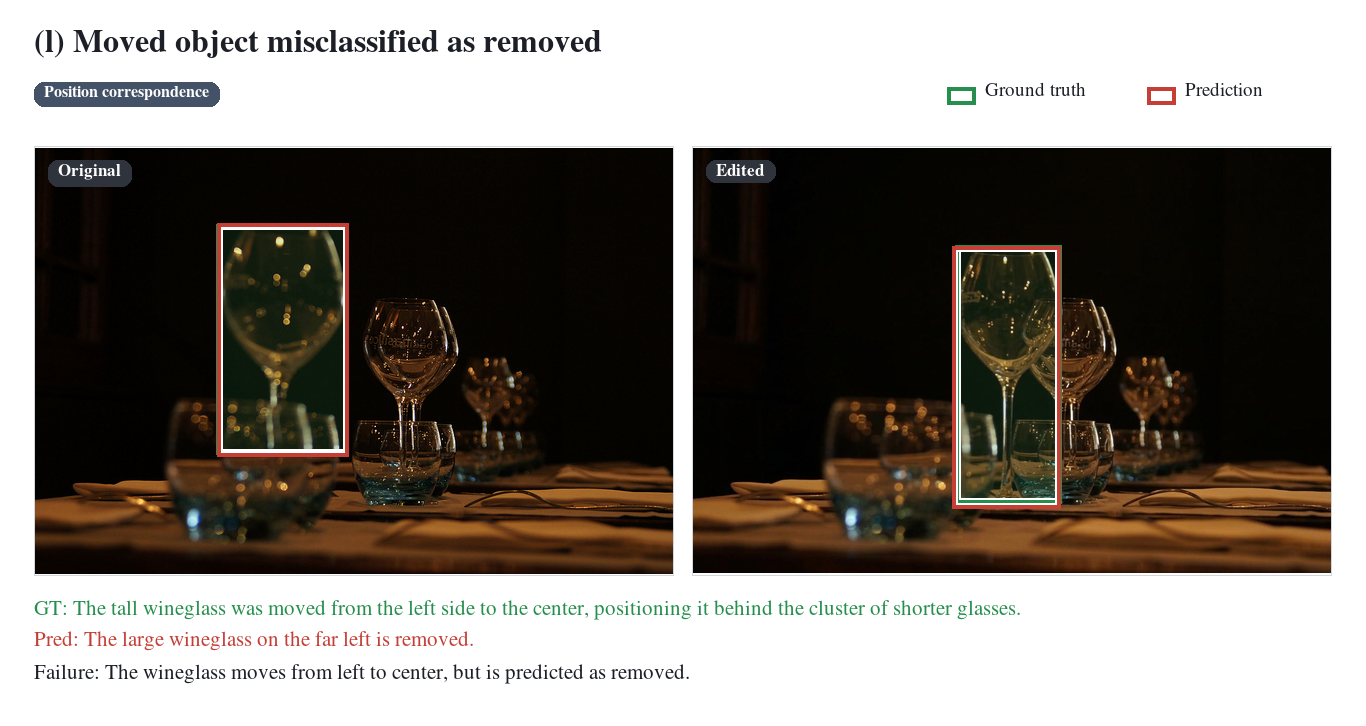}
    \end{subfigure}
    \caption{Typical failure modes of our approach on grounded minimal-change understanding. The examples highlight the model's limitations in handling asymmetric mappings, individuating dense instances and maintaining cross-image position correspondence.}
    \label{fig:failure_cases}
\end{figure}

Despite the improvements brought by SG-ISA, the model still suffers from five recurring failure modes that all stem from weak fine-grained instance binding (as illustrated in Figure~\ref{fig:failure_cases}). These failure modes are broadly consistent with known object hallucination phenomena in large vision-language models~\citep{li2023evaluating,ji2023survey}. First, tiny objects are easily missed, leading to salience-shift hallucinations where the model reports changes in large, salient objects instead (e.g., earring removed $\rightarrow$ ``bouquet removed''). Second, asymmetric changes break the 1-to-1 spatial mapping: the model fails to emit \texttt{<loc\_null>} for additions and falsely triggers it for replacements. Third, dense scenes with multiple similar instances cause counting confusion and overly coarse bounding boxes (e.g., six individual drawers merged into one macroscopic box). Fourth, position changes fail when the model cannot maintain cross-image feature constancy, misclassifying movements as replacements or removals. Finally, distributed attribute changes produce imprecise union boxes instead of tightly localized regions. 

These errors suggest that the bottleneck lies not in scene understanding but in fine-grained spatial correspondence and instance-level grounding. Stronger fine-grained grounding supervision and more explicit cross-image instance binding mechanisms are promising directions for future work.

\section{Attention Visualization}

\begin{figure}[!htbp] %
\centering
\setlength{\tabcolsep}{0pt} %

\begin{subfigure}[b]{0.7\linewidth}
  \centering
  \includegraphics[width=\linewidth]{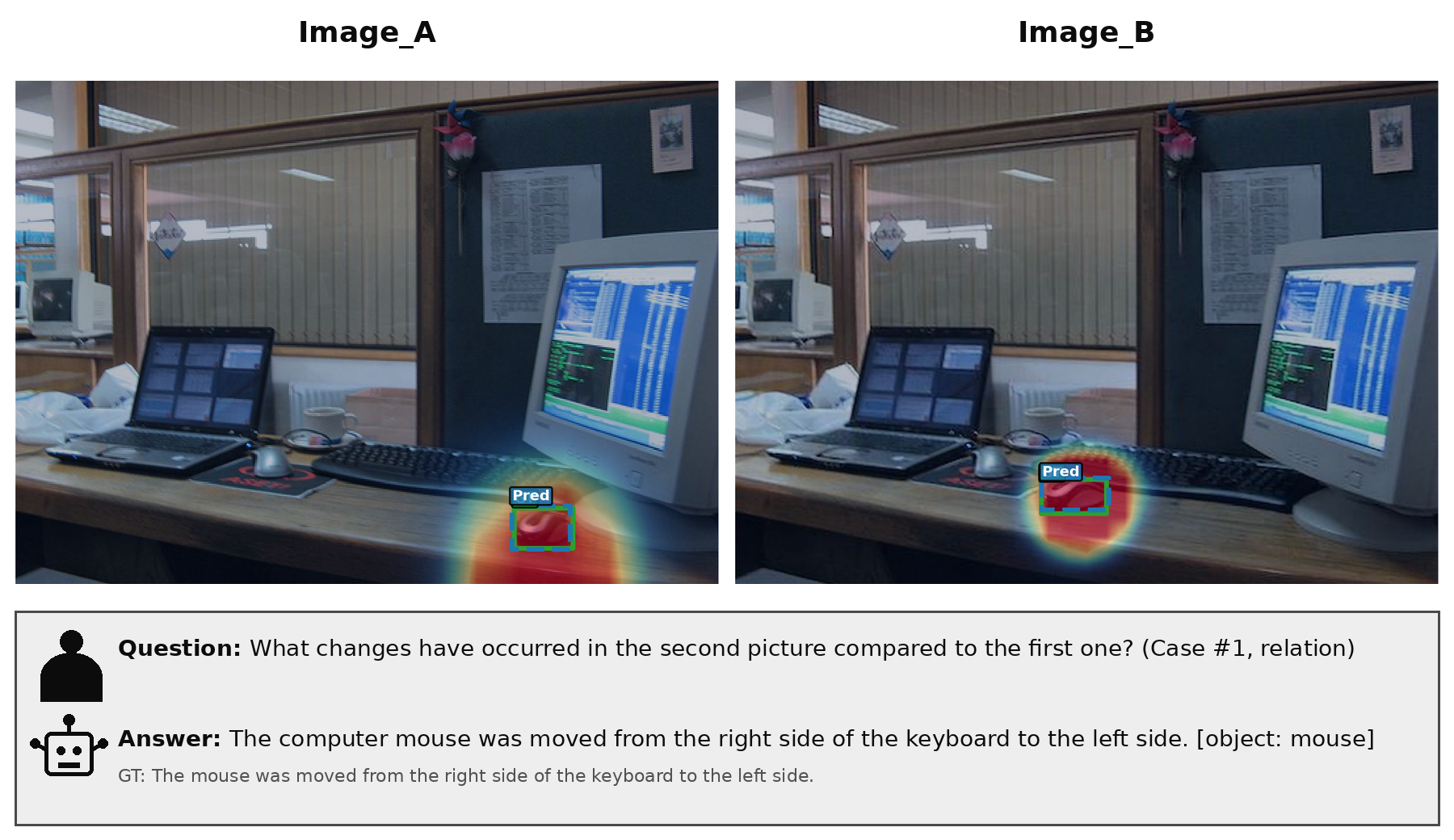}
  \caption{Relocation}
  \label{fig:attn:reloc}
\end{subfigure}
\vspace{0.8em} %
\begin{subfigure}[b]{0.7\linewidth}
  \centering
  \includegraphics[width=\linewidth]{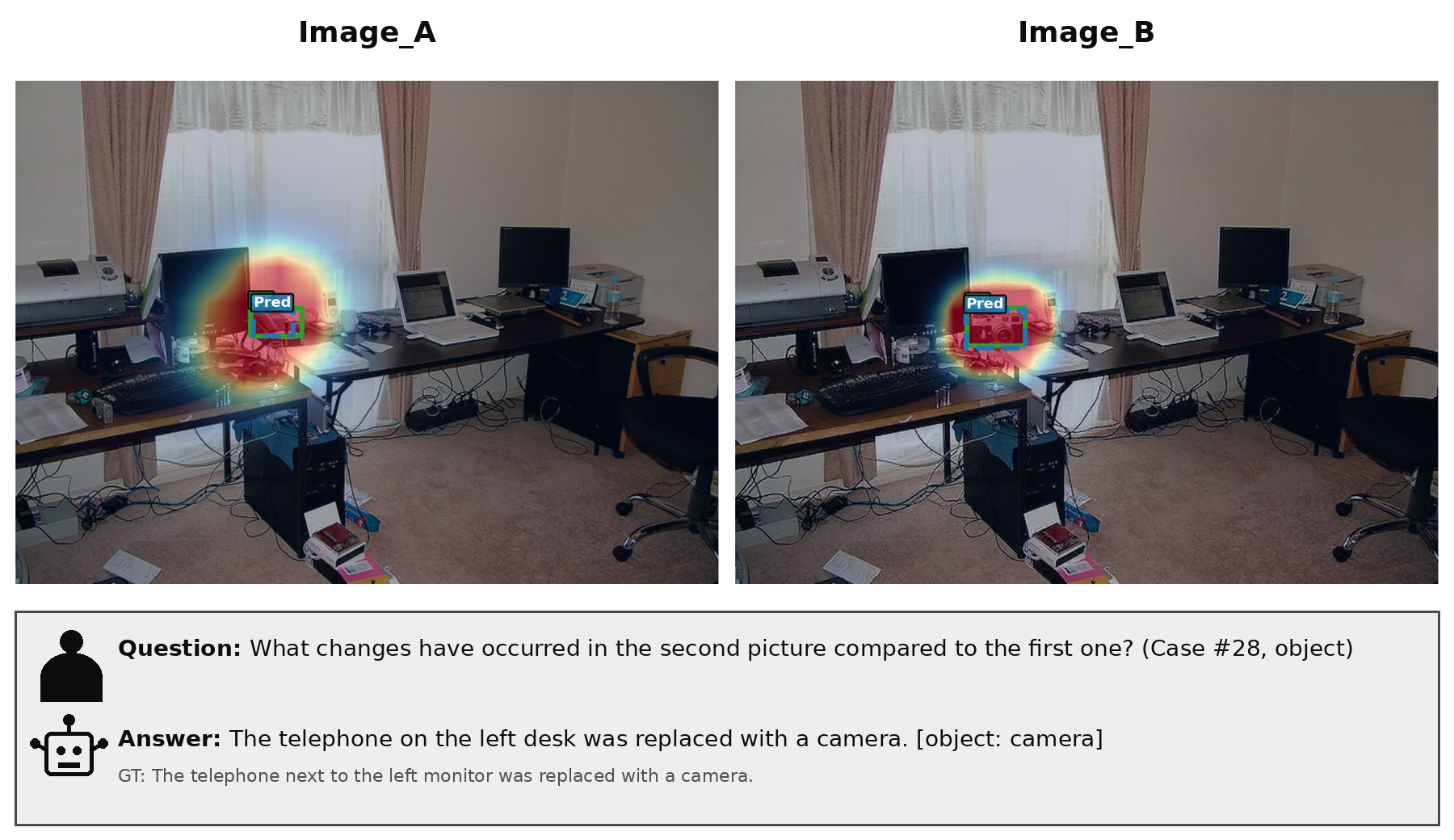}
  \caption{Replacement}
  \label{fig:attn:replace}
\end{subfigure}
\vspace{0.8em}
\begin{subfigure}[b]{0.7\linewidth}
  \centering
  \includegraphics[width=\linewidth]{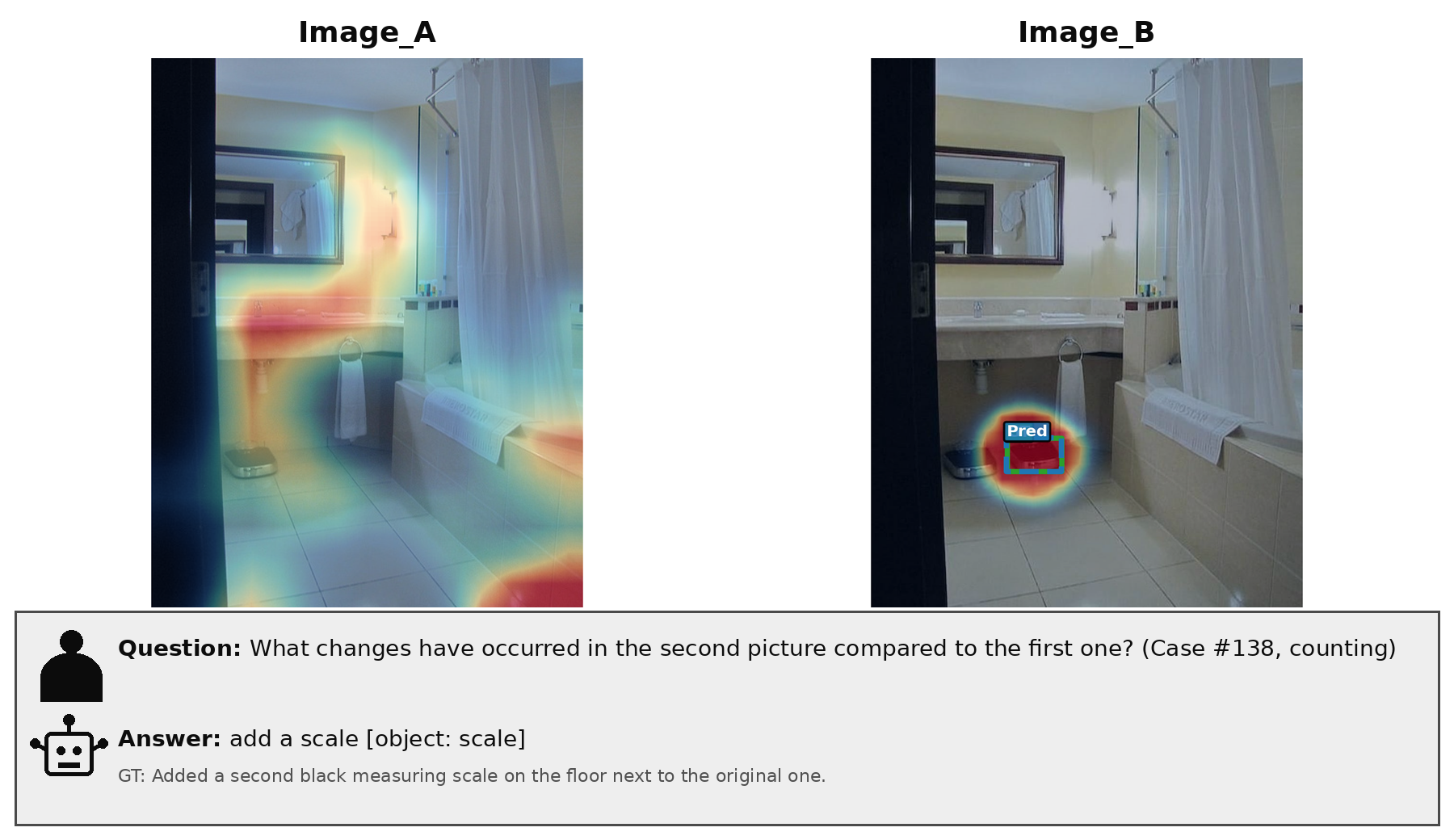}
  \caption{Addition}
  \label{fig:attn:add}
\end{subfigure}

\vspace{0.5em} %
\caption{TAM attention maps across three change types. The model grounds the manipulated object in both images, with sharper activation on the changed image.}
\label{fig:attention}
\end{figure}

We visualize the Token Attention Modulation (TAM) activation maps to inspect how the model locates the target regions. Each panel in Figure~\ref{fig:attention} shows Image A (left) and Image B (right), overlaid with attention heatmaps where red/yellow indicates high activation.

For \textbf{relocation} (Figure~\ref{fig:attn:reloc}), attention forms two distinct hotspots, with one on the object's source position in Image A and another on its target position in Image B, suggesting the model tracks instance identity across spatial displacement. Across all cases, we observe a consistent asymmetry: attention is typically sharper on Image B than on Image A, reflecting that the model uses Image B as the primary signal source while treating Image A as a reference for cross-image discrepancy.

\section{Robustness to Image Sequence Shuffling}

A potential concern when training MLLMs on dual-image inputs is that the model might overfit to the fixed input sequence (i.e., Image A is always the reference and Image B is always the target) rather than genuinely learning to compare visual content. If the model relies primarily on such positional biases or sequential templates, its performance should severely collapse when the input order of the two images is swapped, as the visual changes would no longer align with the memorized sequence prior.

To evaluate robustness against such sequence biases and verify true generalization, we construct a shuffled variant of the \ours~benchmark by randomly swapping Image A and Image B for 50\% of the 2,000 test pairs. Critically, when a pair is swapped, the corresponding bounding-box annotations and semantic entity labels are dynamically updated to maintain spatial correctness, while the natural-language difference captions remain unchanged (as they describe the mutual semantic change between the pair). We then evaluate our SG-ISA model (trained exclusively on the standard, fixed-order training set) on this shuffled benchmark. The evaluation reports only localization metrics, since the captioning objective is unaffected by the shuffling.

Table~\ref{tab:shuffle_results} compares the localization performance of SG-ISA on the shuffled benchmark against the original SFT baseline and SG-ISA, along with representative closed-source models on the original benchmark.

\begin{table}[!htbp]
  \caption{Localization metrics under image sequence shuffling. Gr-F$_1$: Grounded F$_1$; mIoU: mean Intersection over Union; mAP: COCO-style mean Average Precision~\citep{lin2014coco}; HR: hallucination rate ($\downarrow$). \textbf{Bold}: best overall; \underline{underlined}: second best.}
  \label{tab:shuffle_results}
  \centering
  \small
  \setlength{\tabcolsep}{4pt}
  \begin{tabular}{lcccc}
    \toprule
    Model & Gr-F$_1$ ($\uparrow$) & mIoU ($\uparrow$) & mAP ($\uparrow$) & HR ($\downarrow$) \\
    \midrule
    \multicolumn{5}{l}{\textit{Original benchmark}} \\
    GPT~5.4~\citep{openai2026gpt54} & 19.58 & 24.35 & ~~1.96 & 89.35 \\
    Gemini~3.1~Pro~\citep{google2026gemini31pro} & 42.89 & 44.97 & 27.98 & 68.85 \\
    SFT Baseline (2B) & 72.75 & 64.41 & 38.10 & ~~7.25 \\
    SG-ISA~(2B) & 82.88 & 72.51 & 48.77 & ~~3.10 \\
    \midrule
    \multicolumn{5}{l}{\textit{Shuffled benchmark (50\% swap)}} \\
    \rowcolor{gray!10}
    SG-ISA~(2B) & 73.88 & 67.56 & 35.89 & 11.45 \\
    \bottomrule
  \end{tabular}
\end{table}

SG-ISA on the shuffled benchmark achieves 73.88 Gr-F$_1$ and 67.56 mIoU, representing a moderate drop from its original 82.88 Gr-F$_1$ and 72.51 mIoU. However, it still substantially outperforms the SFT baseline trained and evaluated on the original, unshuffled benchmark (72.75 Gr-F$_1$, 64.41 mIoU). This preservation of strong localization ability under shuffled conditions indicates that SG-ISA does not merely memorize input sequences or overfit to positional priors. Instead, the model primarily relies on robust cross-image visual comparison to dynamically localize semantic changes.

The per-category breakdown (Table~\ref{tab:shuffle_category}) further supports this interpretation. Attribute and Object changes, which involve structurally symmetric visual modifications (e.g., color swaps or entity substitutions), remain highly robust under shuffling (86.94 and 77.45 Gr-F$_1$, respectively). In contrast, Counting changes, which are inherently directional and require precise instance-level reasoning (e.g., distinguishing the appearance vs. disappearance of an object), degrade more sharply (55.63 Gr-F$_1$) and exhibit a higher hallucination rate (35.40\%). This asymmetry aligns with the intuition that counting under shuffled conditions amplifies spatial confusion, as the model must dynamically resolve the logical direction of the change without relying on a fixed reference-target sequence.

\begin{table}[!htbp]
  \caption{Per-category localization metrics under shuffling, including Gr-F$_1$, mIoU, COCO-style mAP~\citep{lin2014coco}, and hallucination rate (HR; $\downarrow$).}
  \label{tab:shuffle_category}
  \centering
  \small
  \setlength{\tabcolsep}{4pt}
  \begin{tabular}{lccccc}
    \toprule
    Category & Gr-F$_1$ ($\uparrow$) & mIoU ($\uparrow$) & mAP ($\uparrow$) & AP@50 ($\uparrow$) & HR ($\downarrow$) \\
    \midrule
    Attribute & 86.94 & 80.14 & 56.26 & 76.22 & ~~1.40 \\
    Object    & 77.45 & 68.96 & 40.65 & 63.02 & ~~7.60 \\
    Relation  & 71.62 & 61.59 & 30.84 & 53.03 & ~~1.40 \\
    Counting  & 55.63 & 59.56 & 17.91 & 31.74 & 35.40 \\
    \midrule
    Overall   & 73.88 & 67.56 & 35.89 & 55.79 & 11.45 \\
    \bottomrule
  \end{tabular}
\end{table}

Taken together, these results demonstrate that while SG-ISA may absorb a minor sequential bias during training, the dominant driver of its strong localization performance is genuine fine-grained visual comparison rather than superficial pattern matching. The shuffled benchmark serves as a diagnostic tool confirming that our model reliably learns \emph{how the images differ} rather than just memorizing a fixed input distribution, proving its strong generalization capabilities.

\section{Zero-Shot Generalization to Real-World Image Pairs}

Building upon the robust sequence-invariance demonstrated above, a further critical test of our model's capability is whether its fine-grained cross-image reasoning generalizes to entirely different visual distributions unseen during training. To evaluate this, we take the SG-ISA model trained solely on our \ours~training set and evaluate it zero-shot on a held-out dataset. This dataset comprises 501 manually annotated image pairs (46 Attribute, 30 Counting, 214 Object, and 211 Relation changes) collected from natural photographs.

Table~\ref{tab:real_world_generalization} reports the zero-shot performance of both the base model (pre-trained on generic vision-language tasks but without \ours~data) and our SG-ISA model. The base model achieves 18.43 Gr-F$_1$ and 32.58 mIoU overall, coupled with a low mAP (8.70) and poor caption quality (CIDEr 9.79, SPICE 0.00). This pattern of weak localization and semantically vacuous descriptions is characteristic of models that have not been explicitly aligned for cross-image spatial comparison.

\begin{table}[!htbp]
  \caption{Zero-shot generalization to real-world image pairs. Base: pre-trained model without \ours~fine-tuning. SG-ISA: trained on our \ours~set only. Evaluation includes Gr-F$_1$, mIoU, COCO-style mAP~\citep{lin2014coco}, CIDEr~\citep{vedantam2015cider}, and SPICE~\citep{anderson2016spice}. \textbf{Bold}: best in column; \underline{underlined}: second best.}
  \label{tab:real_world_generalization}
  \centering
  \small
  \setlength{\tabcolsep}{4pt}
  \begin{tabular}{lccccc}
    \toprule
    Model & Gr-F$_1$ ($\uparrow$) & mIoU ($\uparrow$) & mAP ($\uparrow$) & CIDEr ($\uparrow$) & SPICE ($\uparrow$) \\
    \midrule
    \multicolumn{6}{l}{\textit{Overall (501 pairs)}} \\
    Base (Zero-shot) & 18.43 & 32.58 & ~~8.70 & ~~9.79 & ~~0.00 \\
    SG-ISA~(Ours) & \textbf{32.54} & \textbf{56.84} & \textbf{32.65} & \textbf{43.31} & \textbf{20.09} \\
    \midrule
    \multicolumn{6}{l}{\textit{Attribute (46 pairs)}} \\
    Base (Zero-shot) & \textbf{22.50} & 47.28 & 20.99 & ~~3.11 & \textbf{13.93} \\
    SG-ISA~(Ours) & 21.98 & \textbf{50.70} & \textbf{28.38} & \textbf{54.35} & 13.73 \\
    \midrule
    \multicolumn{6}{l}{\textit{Counting (30 pairs)}} \\
    Base (Zero-shot) & 34.78 & 40.95 & 25.30 & ~~4.04 & 27.41 \\
    SG-ISA~(Ours) & \textbf{40.74} & \textbf{69.82} & \textbf{61.48} & \textbf{94.40} & \textbf{29.75} \\
    \midrule
    \multicolumn{6}{l}{\textit{Object (214 pairs)}} \\
    Base (Zero-shot) & \textbf{14.37} & 32.69 & 12.19 & 11.81 & ~~0.00 \\
    SG-ISA~(Ours) & 12.86 & \textbf{58.75} & \textbf{35.36} & \textbf{42.65} & \textbf{20.97} \\
    \midrule
    \multicolumn{6}{l}{\textit{Relation (211 pairs)}} \\
    Base (Zero-shot) & 19.11 & 28.06 & ~~4.57 & ~~9.72 & \textbf{19.72} \\
    SG-ISA~(Ours) & \textbf{47.90} & \textbf{54.38} & \textbf{31.23} & \textbf{34.10} & 19.21 \\
    \bottomrule
  \end{tabular}
\end{table}

SG-ISA substantially outperforms the base model across nearly all metrics. On the overall dataset, SG-ISA improves Gr-F$_1$ by 76\% relatively (18.43 $\to$ 32.54) and mIoU by 74\% relatively (32.58 $\to$ 56.84). The mAP gap is even more pronounced: 32.65 vs.~8.70, a 3.75$\times$ improvement, indicating that our training paradigm equips the model with a highly transferable ability to both detect and localize visual changes in entirely new domains.The text generation metrics tell an equally compelling story. On the overall benchmark, SG-ISA achieves a CIDEr score of 43.31 compared to the base model's 9.79, a 4.4$\times$ improvement that demonstrating that SG-ISA produces descriptions that are far more consistent with human annotations in terms of relevance and specificity. The SPICE score rises from 0.00 to 20.09, confirming that while the base model fails to capture meaningful semantic propositions regarding the changes, SG-ISA successfully articulates structured scene modifications.

The per-category breakdown reveals an informative pattern. Relation changes show the largest improvement: SG-ISA achieves 47.90 Gr-F$_1$ and 54.38 mIoU, more than doubling the base model's performance. This suggests that spatial-relationship reasoning, a core structural skill emphasized by SG-ISA's implicit anchors, transfers robustly across varying visual domains. Counting changes also exhibit dramatic gains across all metrics, including a 23$\times$ improvement in CIDEr (94.40 vs.~4.04), reflecting the model's adeptness at enumerating instance-level variations. Object changes show the most striking SPICE improvement (0.00 $\to$ 20.97), indicating that the base model's vocabulary for object-level differences is largely uninformative, whereas SG-ISA reliably describes additions and removals.

Attribute changes present a more nuanced picture: the base model achieves a marginally higher Gr-F$_1$ (22.50 vs.~21.98), but SG-ISA dominates in mIoU (50.70 vs.~47.28), mAP (28.38 vs.~20.99), and especially CIDEr (54.35 vs.~3.11). The near-zero CIDEr for the base model on Attribute tasks implies that its descriptions remain generic, while SG-ISA generates highly specific, reference-aligned captions.

Taken together, these results confirm that SG-ISA's core capability is not overfitted to the specific visual distribution of the training set. The fact that structural categories (Relation and Counting) show the strongest transfer, while low-level appearance matching (Attribute) is inherently more susceptible to domain shifts, aligns with the intuition that geometric spatial reasoning generalizes better than pixel-level texture mappings. This real-world evaluation provides converging evidence that SG-ISA acquires a domain-agnostic, robust visual comparison mechanism.

\section{Additional Visualization Results}
\label{sec:appendix_additional_vis}

\begin{figure}[thbp]
\centering
\includegraphics[width=\linewidth]{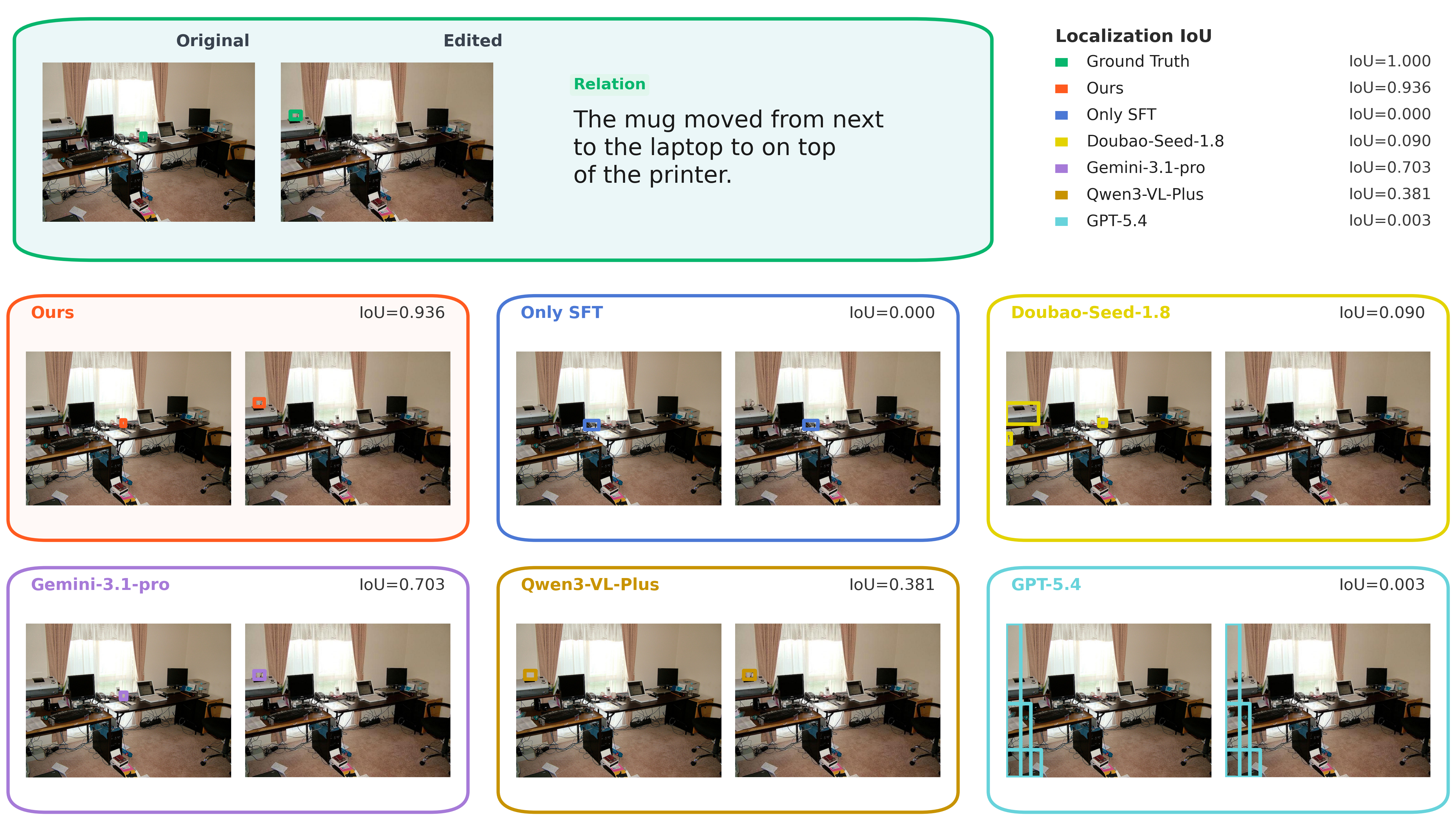}

\vspace{0.6em}

\includegraphics[width=\linewidth]{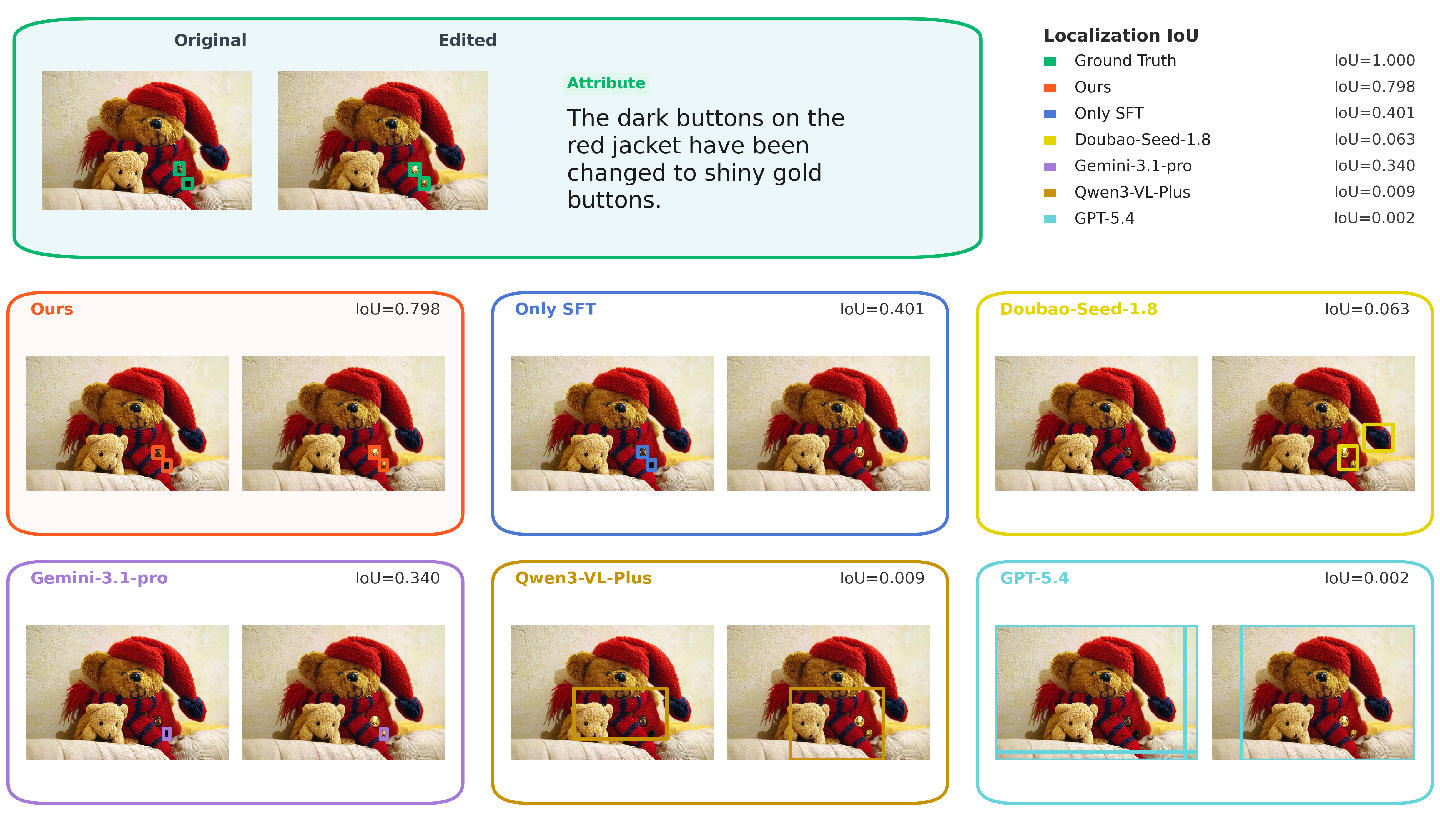}

\caption{Additional qualitative results of SG-ISA on the \ours{} benchmark. Each panel shows the input image pair (Image A and Image B), the model's predicted bounding boxes, and the generated change description, demonstrating the model's capability across diverse change categories and difficulty levels.}
\label{fig:additional_vis_1}
\end{figure}

\begin{figure}[thbp]
\centering

\includegraphics[width=\linewidth]{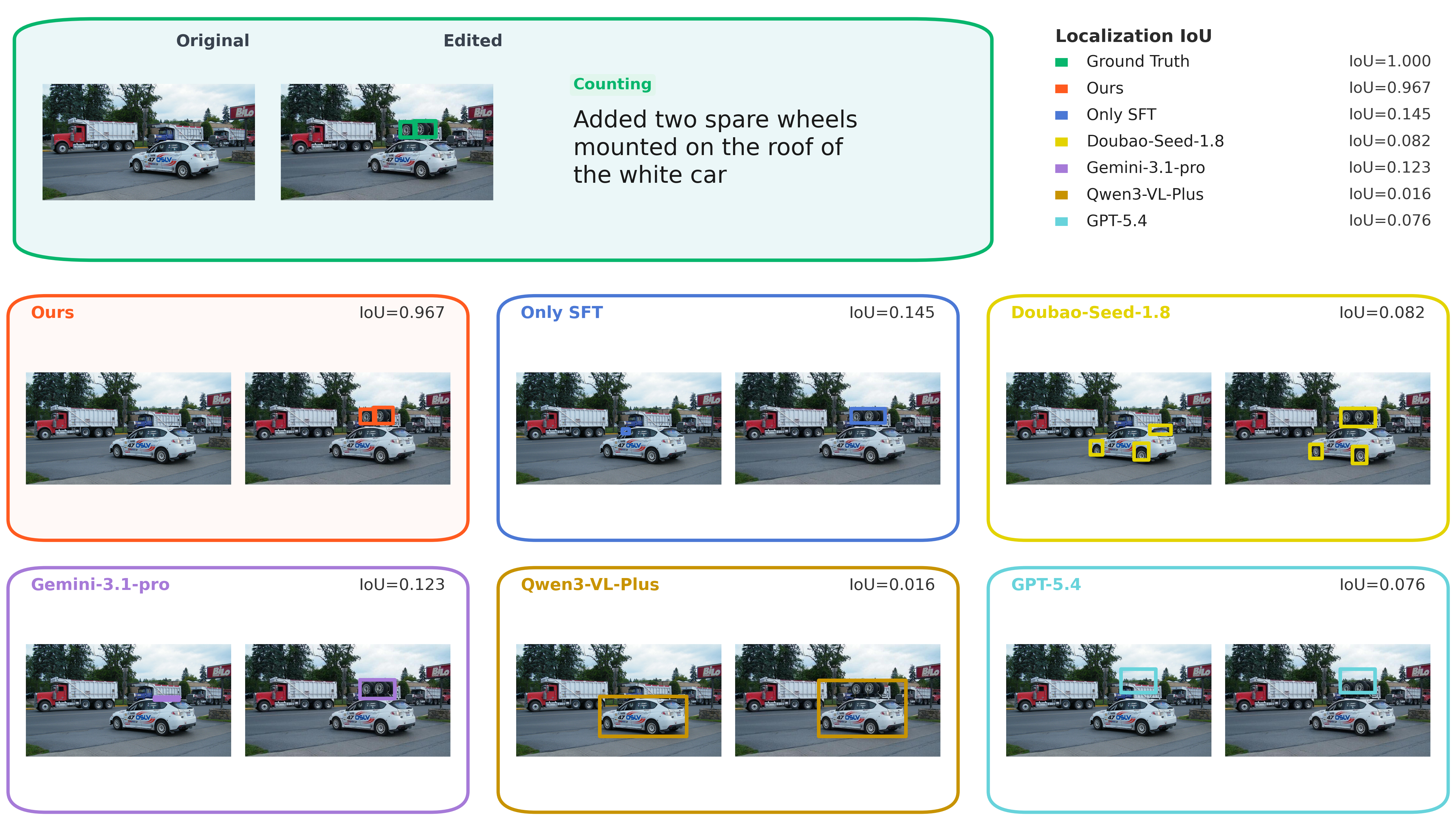}

\vspace{0.6em}

\includegraphics[width=\linewidth]{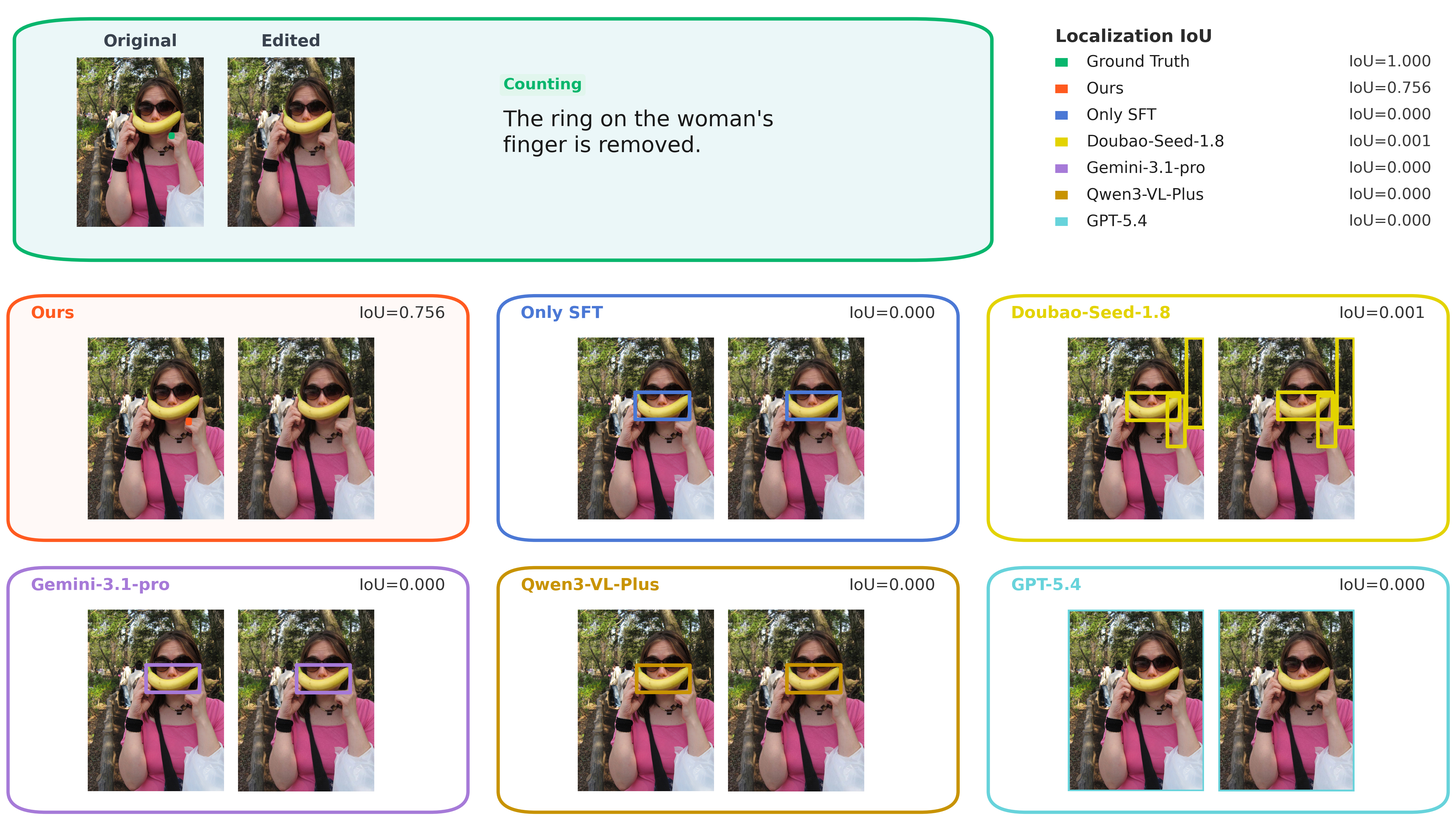}

\caption{Additional qualitative results of SG-ISA on the \ours{} benchmark. Each panel shows the input image pair (Image A and Image B), the model's predicted bounding boxes, and the generated change description, demonstrating the model's capability across diverse change categories and difficulty levels.}
\label{fig:additional_vis_2}
\end{figure}

Figure~\ref{fig:additional_vis_1} and~\ref{fig:additional_vis_2} presents additional qualitative results of SG-ISA across the four change categories, covering a diverse range of scenarios including attribute modifications, object substitutions, counting variations, and spatial relocations. Each example shows the input image pair along with the model's predicted bounding boxes and generated descriptions.

\section{Licenses of Assets Used}
\label{sec:appendix_licenses}

In this work, we utilize several open-source datasets, models, and codebases. The annotations and official website content of the LVIS dataset~\citep{gupta2019lvis} are licensed under the Creative Commons Attribution 4.0 International License (CC BY 4.0). For our models, we employ the Qwen3-VL model~\citep{bai2025qwen3vl}. Both the official model weights and code are released under the Apache License 2.0. We strictly adhere to the terms of use and attribution requirements specified by these respective licenses.

\section{Detailed Definitions of Change Categories}
\label{sec:appendix_category_defs}

In Section 3, we briefly outline the four atomic change categories comprising the \ours{} benchmark. Here, we provide a more comprehensive definition of each category, detailing the specific variations they entail and how they challenge multimodal visual reasoning.

\textbf{Attribute:} 
The Attribute category encompasses fine-grained modifications to the visual properties of an existing entity without altering its core identity. This includes changes in color (e.g., a green shirt changed to blue), material, texture, shape, or physical state. Recognizing these variations requires the model to tightly align the bounding box across the image pair while discerning pixel-level appearance shifts, emphasizing exact descriptive vocabulary matching over spatial shifting.

\textbf{Object:} 
The Object category involves the semantic alteration of instances within the scene. This primarily focuses on the replacement operation, where an existing object is substituted with one of a different category (e.g., swapping a coffee mug for a teacup). To succeed, the model must exhibit a robust understanding of object semantics, accurately recognizing the category shift and updating the entity label while maintaining the contextual awareness of the surrounding environment.

\textbf{Counting:} 
The Counting category manipulates the absolute number of instances of a specific object class. For example, the number of apples on a plate might increase from two to four. This change type is inherently challenging because it demands precise instance-level individuation rather than holistic scene understanding. The model must accurately discern and enumerate identical items, pinpointing exactly which specific instances were newly added or removed amidst a structurally similar dense background.

\textbf{Relation:} 
The Relation category evaluates spatial dynamism and structural understanding. In these pairs, an entity's semantic identity and appearance are preserved, but its physical position or interaction with the environment is altered (e.g., moving a vase from the left side of a table to the right). To succeed, the model must establish robust cross-image object correspondence and track geometric displacement, proving it grasps relative spatial transformations rather than merely detecting static object presence.

\end{document}